\documentclass[sigconf, screen]{acmart}

\usepackage{multirow}
\usepackage{colortbl}
\usepackage{subfigure}

\AtBeginDocument{
  }

\copyrightyear{2025}
\acmYear{2025}
\setcopyright{cc}
\setcctype{by-nc}
\acmConference[MM '25]{Proceedings of the 33rd ACM International Conference on Multimedia}{October 27--31, 2025}{Dublin, Ireland}
\acmBooktitle{Proceedings of the 33rd ACM International Conference on Multimedia (MM '25), October 27--31, 2025, Dublin, Ireland}
\acmDOI{10.1145/3746027.3754553}
\acmISBN{979-8-4007-2035-2/2025/10}

\begin{document}

\title{Toxicity Begets Toxicity: Unraveling Conversational Chains in Political Podcasts}

\author{Naquee Rizwan}
\affiliation{%
  \institution{Indian Institute of Technology}
  \city{Kharagpur}
  \state{West Bengal}
  \country{India}}
\email{nrizwan@kgpian.iitkgp.ac.in}
\orcid{0009-0007-1872-6618}

\author{Nayandeep Deb}
\affiliation{%
  \institution{Indian Institute of Technology}
  \city{Kharagpur}
  \state{West Bengal}
  \country{India}}
\email{nayandeepdeb125@kgpian.iitkgp.ac.in}
\orcid{0009-0001-0828-2511}

\author{Sarthak Roy}
\affiliation{
  \institution{Indian Institute of Technology}
  \city{Kharagpur}
  \state{West Bengal}
  \country{India}}
\email{sarthak.cse22@kgpian.iitkgp.ac.in}
\orcid{0009-0007-1060-2266}

\author{Vishwajeet Singh Solanki}
\affiliation{
  \institution{Indian Institute of Technology}
  \city{Kharagpur}
  \state{West Bengal}
  \country{India}
}
\email{vsinghsolanki@kgpian.iitkgp.ac.in}
\orcid{0009-0001-9257-9130}

\author{Kiran Garimella}
\affiliation{
 \institution{Rutgers University}
 \city{New Brunswick}
 \state{New Jersey}
 \country{USA}}
\email{kiran.garimella@rutgers.edu}
\orcid{0000-0002-0173-5557}

\author{Animesh Mukherjee}
\affiliation{
  \institution{Indian Institute of Technology}
  \city{Kharagpur}
  \state{West Bengal}
  \country{India}}
\email{animeshm@cse.iitkgp.ac.in}
\orcid{0000-0003-4534-0044}

\renewcommand{\shortauthors}{Naquee Rizwan et al.}

\newcommand{\sysGPT}[1]{\textsc{GPT-4o}}
\newcommand{\sysGPTA}[1]{\textsc{GPT-4o-Audio}}

\newcommand{\sysQWEN}[1]{\textsc{Qwen-2}}
\newcommand{\sysQWENA}[1]{\textsc{Qwen-2-Audio}}

\begin{abstract}
Tackling toxic behavior in digital communication continues to be a pressing concern for both academics and industry professionals. While significant research has explored toxicity on platforms like social networks and discussion boards, podcasts—despite their rapid rise in popularity—remain relatively understudied in this context. This work seeks to fill that gap by curating a dataset of political podcast transcripts and analyzing them with a focus on conversational structure. Specifically, we investigate how toxicity surfaces and intensifies through sequences of replies within these dialogues, shedding light on the organic patterns by which harmful language can escalate across conversational turns.
\textit{\textbf{\textcolor{red}{Warning: Contains potentially abusive/toxic contents.}}}\footnote{\textcolor{red}{Accepted as regular paper \textit{(main track)} at ACM Multimedia 2025}}
\end{abstract}



\maketitle

% ------------------------------------------------------------

\section{Introduction}

Understanding and addressing toxicity in digital media is an ongoing challenge for researchers and practitioners alike. While much attention has been paid to platforms such as social media and online forums, comparatively less is known about the nature and dynamics of toxicity within the rapidly growing medium of podcasts. In this study, we aim to bridge this gap by collecting and analyzing political podcast data, focusing specifically on conversation chains -- structured reply patterns within podcast transcripts -- to study the emergence and propagation of toxicity in this medium, naturally unfolding how \textit{toxicity begets toxicity} through conversation chains.

\noindent\textbf{Rise in the popularity of podcasts}: Podcasting has seen remarkable growth in recent years, emerging as a mainstream medium for information and entertainment. As of 2024, there are over 500 million podcast listeners worldwide, representing 23.5\% of all internet users. In the United States alone, 47\% of adults have listened to a podcast within the last month, a figure that has more than tripled over the past decade. Politics and government rank as the third most popular podcast topic, with 41\% of listeners regularly tuning in to content in this category\footnote{\url{https://www.statista.com/statistics/270365/audio-podcast-consumption-in-the-us/}\\\url{https://www.edisonresearch.com/the-podcast-consumer-2024-by-edison-research/}}. This growth underscores the importance of podcasts as a platform for political discourse, but it also highlights their potential as vehicles for misinformation, conspiracies, and hate speech~\cite{wirtschafter2023audible}. Unlike traditional forms of media, podcasts remain largely un-moderated, further exacerbating the risks of unchecked toxicity. Understanding toxicity in podcasts is therefore not only academically significant but also crucial for ensuring healthy democratic discourse.

\noindent\textbf{Need of the hour:} A recent post by Meta\footnote{\url{https://about.fb.com/news/2025/01/meta-more-speech-fewer-mistakes/}} highlights the difficulty of real-time proactive content moderation and one such massive outrage that recently broke out in India\footnote{\url{https://www.ndtv.com/feature/indias-got-latent-row-who-said-what-as-samay-raina-forced-to-remove-all-youtube-episodes-7708513}} is a prime example of the need of the hour. Studying toxicity in podcasts presents unique challenges thus limiting the effectiveness of real-time proactive content moderation. Unlike one-off contents like text-based posts, memes, and GIFs, the audio format of podcasts makes them inherently difficult to monitor and analyze using traditional strategies, primarily due to their continuous nature. Transcribing large volumes of audio content incurs significant computational and financial costs, making large-scale studies impractical without significant resources. Further, audience interaction in podcasts is limited compared to social media platforms, where users can directly respond to or fact-check content. This lack of interactivity complicates efforts to understand how toxic narratives resonate with audiences. Moreover, analyzing toxicity within podcast transcripts requires not only identifying toxic language but also tracking how it ebbs and flows over time -- examining what triggers toxic conversations and how they evolve within conversational structures. These complexities highlight why \textbf{na\"ive approaches to understand toxicity in podcasts often fail} to capture its multifaceted nature.
Existing research provides valuable starting points, but fails to address the problem in a comprehensive way. Advances in automatic transcription tools, such as Whisper, now allow for large-scale analysis of audio data. However, as previous work on datasets like RadioTalk~\cite{beeferman2019radiotalk} demonstrates, transcription errors and content ambiguity remain significant barriers, particularly for conversational and emotive formats such as podcasts. 
In addition, though the study of toxicity online is an active research area~\cite{paz2020hate}, there is little work on understanding conversational structures of toxicity and how it evolves~\cite{zhang2024imperative}.

\noindent\textbf{Contributions:} In this paper, we introduce a \textbf{fresh research statement} for audio content moderation proposing a \textbf{\textit{novel}} concept of toxic conversation chains and utilizing them in the study of how \textit{toxicity begets toxicity}. This formulation allows us to understand the growth, spread and decay of toxicity in conversations. Using state-of-the-art transcription models and conversational analysis techniques, we explore how toxicity emerges, spreads and decays in political podcasts. Our approach combines scalable transcription workflows with innovative analysis of conversational dynamics, providing crucial insights into the structure and flow of toxic discourse in this medium. 
Our work represents a critical step towards systematically analyzing toxicity in podcasts and lays the foundation for future research in this area. Our \textbf{key contributions and observations} are as follows:\\
\noindent\textbf{(i)} Collecting, transcribing and diarizing a dataset over 52 popular political podcasts (31 right- and 21 left-leaning) from the US, reaching tens of millions of people, and identify thousands of instances of toxicity in these podcasts. The dataset encompasses a total of \textbf{12,322 episodes} from these channels (9,166 from right- and 3,156 from left-leaning podcasts).\\
\noindent\textbf{(ii)} Identifying concerning trends of a majority of episodes from many popular podcasts containing at least one instance of toxicity.\\
\noindent\textbf{(iii)} Introducing \textbf{toxic conversation chains,} built from the transcriptions where each chain has a highly toxic part which we call the \textit{anchor} and a series of preceding and following parts to express the context of the anchor.\\
\noindent\textbf{(iv)} Analysis of these message chains reveals that the anchor text is (a) longer in duration, (b) more repetitive, (c) incorporates figurative languages such as metaphors and hyperboles, and (d) is closely tied to emotions like anger, fury, and annoyance.\\
\noindent\textbf{(v)} Annotating a \textbf{\textit{novel}} dataset based on the premise of \textit{toxicity begetting toxicity} from top 200 toxic conversation chains (100 each from right- and left-leaning channels) to identify change points where toxicity levels shift. Using these annotations, we proposed a setup to predict the temporal trajectory of these conversations. Our findings suggest that such automated methods hold significant potential for real-time monitoring and the development of effective intervention mechanisms in the future\footnote{Supplementary material, dataset and code -- \url{https://github.com/hate-alert/ToxicityBegetsToxicity-Audio}.}.

% ------------------------------------------------------------

\section{Related works}
\textbf{Detection of toxicity}: Defining and detecting toxicity requires nuanced approaches due to its context-sensitive nature. Techniques like BERT-based sentiment analysis have proven effective in detecting toxic content on platforms such as Twitter, leveraging large labeled datasets to improve classification accuracy~\cite{mathew2022hatexplainbenchmarkdatasetexplainable}. Further, context-aware models are necessary to differentiate between benign negativity and harmful toxicity, particularly in multi-dimensional interactions~\cite{gao-huang-2017-detecting} and across different modalities~\cite{rizwan2025exploring}. Perspective API~\cite{lees2022newgenerationperspectiveapi} is a valuable tool for researchers to analyze toxicity of online content, allowing them to study the spread of hate speech, cyberbullying, and other harmful content in different languages.\\
\textbf{Toxicity in podcasts}: Unlike traditional social media, podcasts present unique challenges due to their audio form. Transcription is often necessary for textual analysis, but introduces errors and challenges such as speaker identification and role prediction~\cite{clifton-etal-2020-100000} that can skew results. Podcasts also allow for prolonged discussions, creating more opportunities for toxicity to escalate. Recent work has curated datasets to study toxicity in political podcasts, identifying patterns in conversational structures that promote or mitigate toxicity ~\cite{litterer2024mapping,wirtschafter2021challenge}. Similarly, the authors in~\cite{balci2024podcastoutcastsunderstandingrumbles} analyze the political biases and content strategies of YouTube and Rumble podcasts. As a mitigation strategy, the authors in~\cite{Pathiyan_Cherumanal_2024} propose an approach to combat misinformation in podcasts using auditory alerts to notify listeners of potential misinformation in real-time.\\
\textbf{LLMs and audio-LLMs}: In recent years, the advent of transformers~\cite{vaswani2023attentionneed} has led to the development of various state-of-the-art LLMs~\cite{minaee2024largelanguagemodelssurvey} which have proven their capabilities in very complex applications. However, research on audio-LLMs for conditional generation given an audio input is limited~\cite{chu2024qwen2, openaiGPT4}\footnote{\url{https://platform.openai.com/docs/guides/audio}}. Recently researchers have explored LLMs for change point detection as well~\cite{dong2024llmsservetimeseries}. However, to the best of our knowledge, audio-LLMs have not been employed for such a complex task of detecting change points in a toxic conversation before.

\noindent\textbf{Present work}: To the best of our knowledge, we are the first to look at toxicity in podcasts and use the conversational aspects in podcasts. Our contributions are \textbf{two} fold: first is to show the prevalence of toxicity in popular political podcasts and the development of methods to look at toxicity in conversations rather than considering them as a one-off instance.

\section{Dataset}
In this section, we discuss the details of curating, transcribing and diarizing the dataset being used for this work. 

\noindent\textbf{Curation}: We construct the dataset based on the channels list curated and shared by ~\cite{wirtschafter2023audible} and also from the following collection hosted at \url{https://politicalpodcastproject.shinyapps.io/dataset/}. They identify popular political podcasts by analyzing and selecting talk shows that discuss politics, policy, current events, or news. We further expand the channels by including related political punditry podcasts from the `you might also like' recommendations, creating a comprehensive sample for analysis. \textit{RSS feeds} and the link to episodes from the above collection are used to download the audio files, and the podcasts span over the years 2022-2023. The full list of \textbf{52 podcast channels} (31 right- and 21 left-leaning), making a total of \textbf{12,322 episodes} (with 9,166 from right- and 3,156 from left-leaning channels) is noted in the Appendix\footnote{Care should be taken in interpreting the results particularly in the context of prevalence of toxicity, since our dataset only represents a small subset of political podcasts.}. Even though the set of podcasts we consider is small, these shows might have disproportionate impacts and people who are in power might disproportionately listen to these. For e.g., a lot of January 6, 2021 US Capitol attack\footnote{\url{https://www.britannica.com/event/January-6-U-S-Capitol-attack}} coordination and top down messages iterate through the podcasts of personalities like Steve Bannon. Hence, the claims made on these podcasts may have particularly consequential implications for broader public opinion and political discourse. So, as a part of this work, we try to understand the toxicity being driven through conversational chains so that such content can be auto-moderated and avoided.

\noindent\textbf{Processing}: We use \textsc{Whisper}~\cite{radford2023robust} to transcribe and \textsc{Pyannote}\footnote{\url{https://pyannote.github.io/}} to diarize and identify individual speakers~\cite{bredin2021end}. We filter out overlapping speech segments to avoid performance degradation in speaker clustering during diarization, which is accomplished
using the same \textsc{Pyannote} model.
The diarization process typically achieves high-quality results with over 90\% accuracy, though some misclassifications do occur. For a detailed analysis of diarization quality, please refer to~\cite{bredin2021end}. Since our study focuses on conversation chains (see Section~\ref{sec:toxic_conversation_chains}), it is crucial to maintain continuous speaker conversations without interruptions. In this context, the 90\% accuracy is sufficient for our analysis, as we are examining conversation chains rather than individual speaker behaviors. When necessary, we outline the specific cleaning approaches used in subsequent sections.

\noindent\textbf{Metadata}: After processing, each episode in the dataset is made up of a number of speaker turn chunks, as returned by \textsc{Pyannote} library. Each chunk contains (a) start time (b) end time (c) unique speaker id, and (d) diarized text. All these metadata correspond to these uneven chunks. As a next step, we further organize our data in the Section~\ref{sec:toxic_conversation_chains} for making conversational chains that are uniformly distributed to carry out our analysis. A detailed summary of the dataset over 52 podcast channels, and some more metadata associated with them, are presented in the Appendix.

% ------------------------------------------------------------

\begin{figure}
    \centering
    \includegraphics[width=0.95\columnwidth]{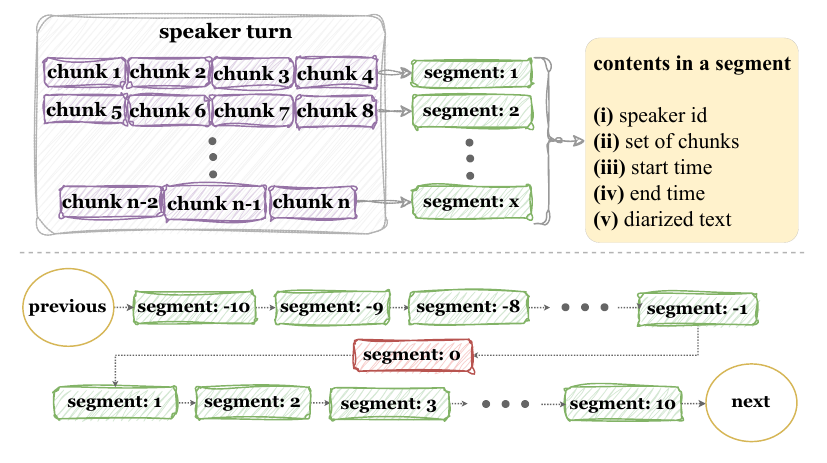}
    \caption{\footnotesize{\textbf{\textsc{Top}}: Computation of segments from chunks and their contents. \textbf{\textsc{Bottom}}: Schema for toxic conversation chains. The segment marked in red color represents the anchored segment with toxicity above a threshold of 0.7.}}
    \label{fig:chain_schema}
\end{figure}

\section{Toxic conversation chains}

% ------------------------------------------------------------

\label{sec:toxic_conversation_chains}
In this section, we discuss the process employed to accurately calculate the toxicity scores and the subsequent formulation of toxic conversation chains using these toxicity scores. Note that our work is centered to understand and intervene upon the trends of toxicity across audio podcasts. Hence, we use \textit{state-of-the-art} toxicity scores generator as an aid to study conversational trends.

\noindent\textbf{Segments}: To compute the toxicity of a speaker's speech we pass the diarized text corresponding to the speech to Perspective API.\footnote{\url{https://developers.perspectiveapi.com/s/about-the-api-attributes-and-languages}} We use the benchmark \textsc{Toxicity} attribute, available with the API to obtain the toxicity scores. Since the Perspective API can handle a limited number of tokens we split the diarized text corresponding to a speaker turn into \textit{chunks} of 17 seconds since this is the median conversation time of a turn in our dataset (see Figure~\ref{fig:chain_schema}).
We club 1-4 chunks (approximately \textit{one} minute) into a \textit{segment}. Each segment corresponds to a specific speaker identified by a speaker ID, a set of (at most) four chunks, the start and end timestamps and the diarized text corresponding to the chunks in the segment. The toxicity score of a segment is set to the maximum of the toxicity score across these chunks within the segment. If a speaker turn exceeds the duration limit of a segment, it is broken into the required number of contiguous segments. All our analyses are at the level of these one minute segments so that the observations are statistically meaningful and not over dominated by a speaker speaking over a long stretch.

\noindent\textbf{Conversation chain}: A toxic chain constitutes an \textit{anchor} segment, the \textit{preceding} \textit{ten} and the \textit{following} \textit{ten} segments of the anchor segment (see Figure~\ref{fig:chain_schema}). A segment is identified as an anchor segment if its toxicity score is \textbf{0.7 or higher}. This threshold is set as per the guidelines noted in documentation of the Perspective API\footnote{\url{https://developers.perspectiveapi.com/s/about-the-api-score}} (refer to \textit{`Supporting human moderators'} subsection). We take the previous and the next \textit{ten} segments to understand what leads to the rise in toxicity and what happens after a peak toxic threshold has been reached. We manually tune to ten previous and next segments, as this gives a perfect amount of back-and-forth context around the anchor point. Choosing fewer segments leads to the loss of context and incorporating more segments was not beneficial. Further, this selection is just precise for the manual CPD annotation as well (refer to Section~\ref{section:cpd}). Anchoring on this threshold, we perform our analysis on a total of \textbf{8,634 chains} from right- and \textbf{7,124 chains} from left-leaning channels. A set of representative examples of toxic conversation chains is provided in Figure~\ref{fig:conversation_main} \& in the Appendix. Further, extended analysis over the properties of these chains (especially for the left-leaning ones) is presented in the Appendix.

\noindent\textbf{Robustness of Perspective API scores}: To establish the robustness of toxicity scores assigned by Perspective API, we perform additional validation. Specifically, we use two different types of models and further verify the anchor segment of toxic conversation chains via human annotation. We perform these additional robustness experiments on the right-leaning toxic conversation chains (results have exactly similar trends for the left-leaning channels and hence are not shown to avoid repetition). Following are the key results:\\
\textbf{(i)} \textsc{Detoxify} \textit{library generated scores}: We run two different \textsc{Toxic-BERT} models (\texttt{original} \& \texttt{unbiased})\footnote{\url{https://dataloop.ai/library/model/unitary_toxic-bert/}} and utilize their \textsc{Toxicity} attribute scores. Both of these models align with nearly \textbf{93\%} of the cases where Perspective API assigns toxicity scores higher than 0.7.\\
\textbf{(ii)} Classification using \sysGPT{}: We prompt \sysGPT{} to classify whether the diarized text of the anchor segment is toxic or not and provide the definition for toxicity from the Perspective API manual as additional context in the prompt. \sysGPT{} classified \textbf{86\%} of the total anchor segments to be toxic aligning with the cases where the toxicity score from Perspective API is greater than 0.7. Please refer to the Appendix for the detailed prompt.\\
\textbf{(iii)} Human evaluation: We conduct binary annotation task (\textit{toxic} or \textit{non-toxic}) on anchor segment's diarized text over 100 randomly selected chains using two expert annotators and found 86 of them to be either toxic, vulgar or attacking in nature to be spoken on a live-streaming show. The results are obtained with a high inter-annotator agreement of 0.69 based on Cohen's $\kappa$. During annotation, annotators were asked to stick to the definition of \textsc{Toxicity} attribute as per the Perspective API documentation.\\
Based on the obtained results, we henceforth conclude that a Perspective API score of 0.7 is a reasonable threshold to flag the anchor text of a chain to be toxic.

\begin{table}[!ht]
\centering
\renewcommand{\arraystretch}{.001}
\begin{tabular}{ll}
\textbf{\includegraphics[width=1\linewidth]{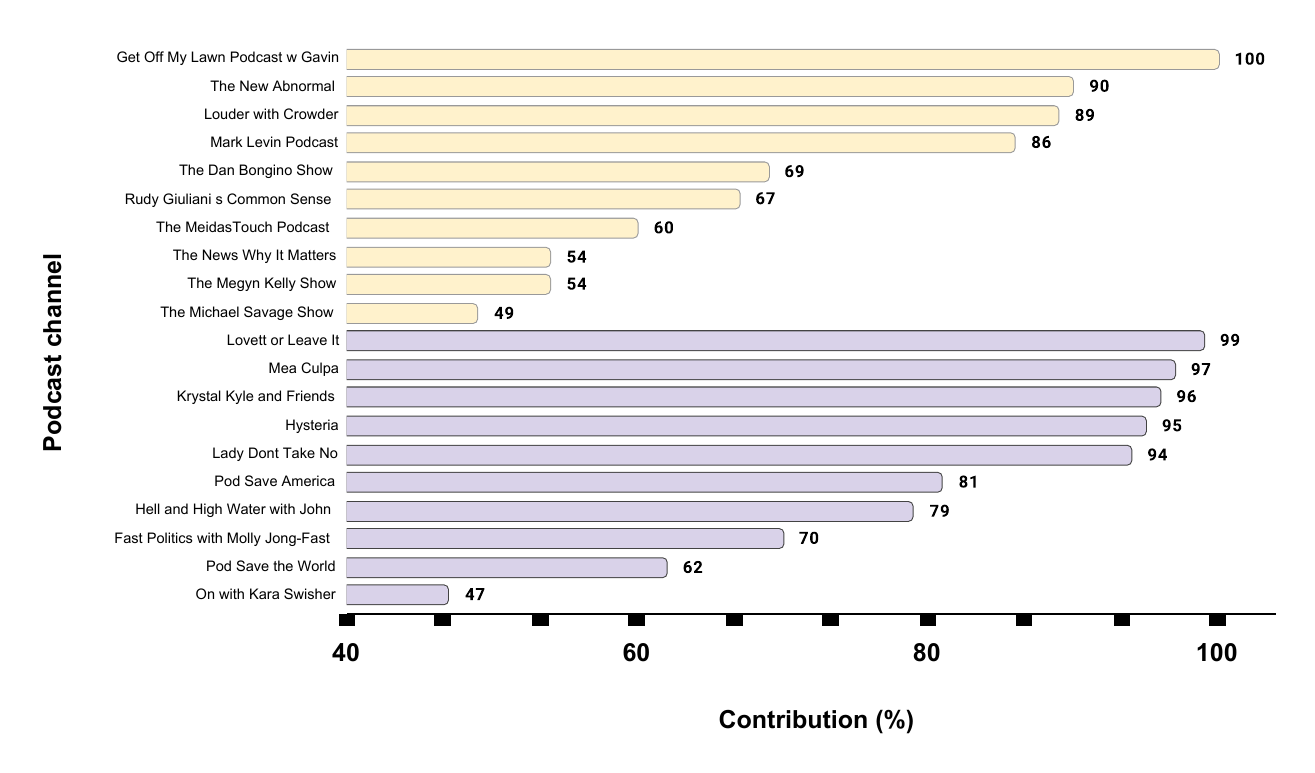}}
\end{tabular}
\captionof{figure}{\label{fig:episodes_distribution} \footnotesize{\footnotesize{\textbf{\textsc{Top 10}} podcast shows with most amount of toxic content for each leaning, i.e. \colorbox[HTML]{fff5c4}{right} and \colorbox[HTML]{d7c7ee}{left}. The bars show the percentage of episodes containing at least one toxic conversation.}}}
\end{table}

\begin{table}[!ht]
\centering
\renewcommand{\arraystretch}{.001}
\begin{tabular}{ll}
\textbf{\includegraphics[width=1\linewidth]{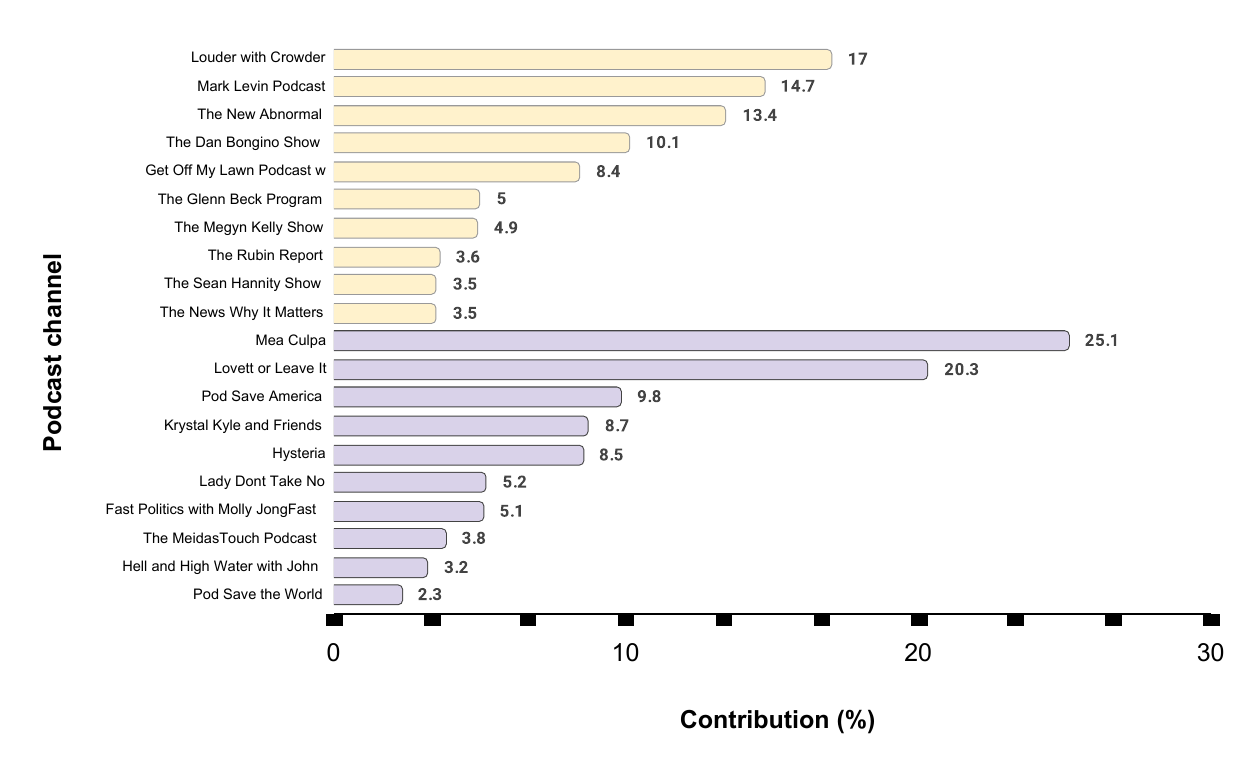}}
\end{tabular}
\captionof{figure}{\label{fig:podcast_distribution} \footnotesize{\textbf{\textsc{Distribution}} of toxic conversation chains across the podcast channels for each leaning, i.e. \colorbox[HTML]{fff5c4}{right} and \colorbox[HTML]{d7c7ee}{left}. Percentage contribution for \textbf{\textsc{Top 10}} podcast channels are shown.}}
\end{table}

\begin{table*}[!ht]
\centering
\begin{tabular}{ll}
\textbf{\includegraphics[width=0.95\linewidth]{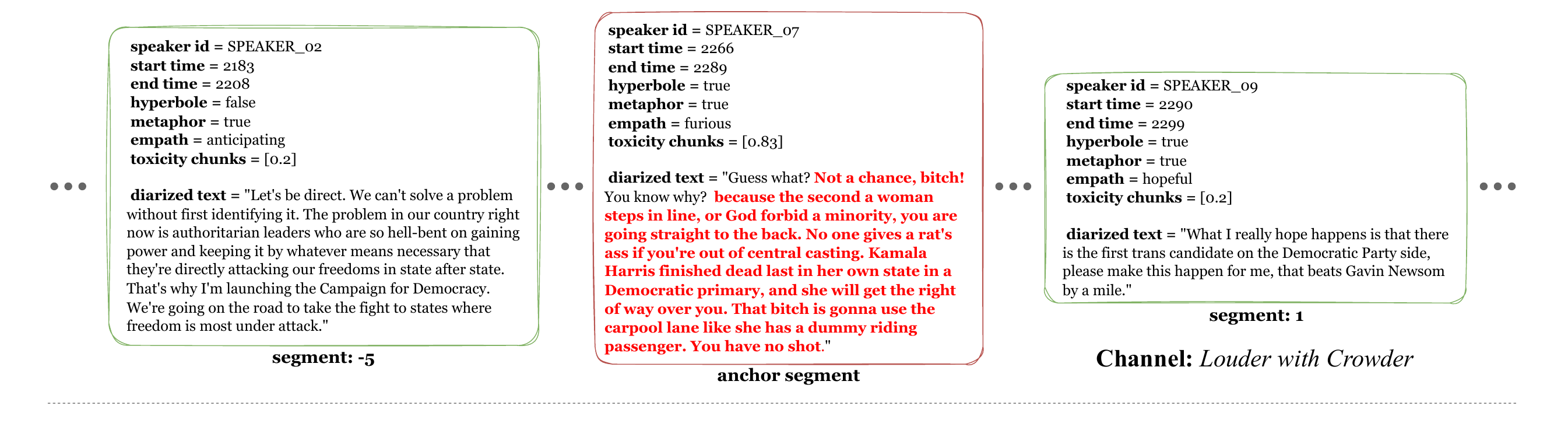}}\\
\textbf{\includegraphics[width=0.95\linewidth]{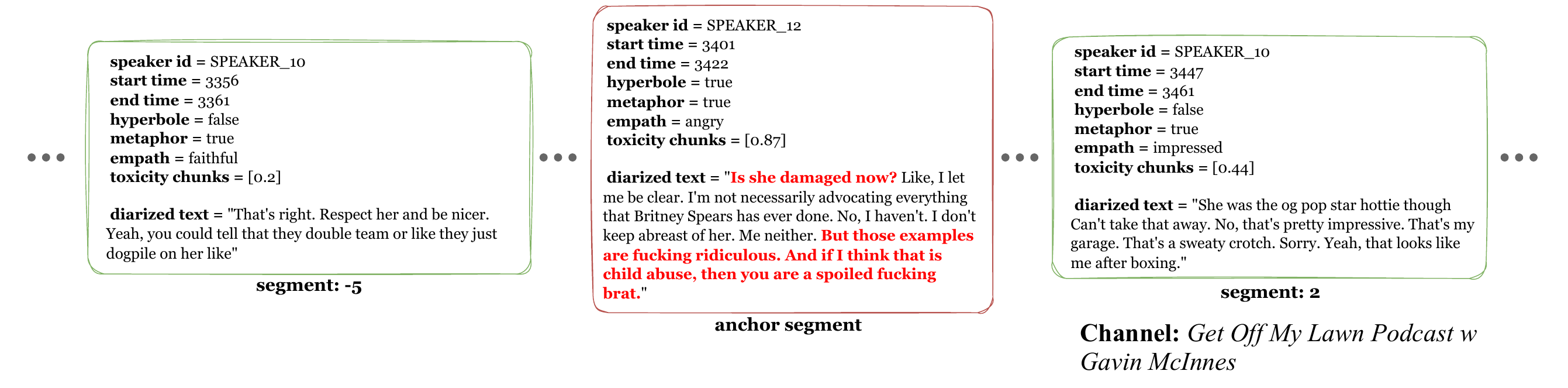}}
\end{tabular}
\captionof{figure}{\label{fig:conversation_main} \footnotesize{\textbf{\textsc{Example}}: Since it is not feasible to illustrate all segments, one among the previous and next segments is shown along with the anchor segment from the toxic conversation chain. Toxic texts in the anchor segment are marked in \textcolor{red}{red}. \textbf{\textsc{Note:}} start and end times are in seconds.}}
\end{table*}

\noindent\textbf{Most toxic channels}: Figure~\ref{fig:episodes_distribution} shows the top 10 podcasts with the highest percentage of episodes containing at least one instance of toxic conversations for both right- and left-leaning channels.
Certain shows such as \textit{Get Off My Lawn Podcast w/ Gavin Mcinnes}, \textit{The New Abnormal}, and \textit{Louder with Crowder} from right-leaning and \textit{Lovett or Leave It}, \textit{Mea Culpa}, \textit{Krystal Kyle and Friends}, \textit{Hysteria} and \textit{Lady Dont Take No} from left-leaning channels, have at least one instance of a toxic conversation in almost all of their episodes. Tables in the Appendix show the full statistics for all the shows in our dataset. It is quite surprising to see instances of toxicity in hundreds of episodes, even on shows with tens of millions of listeners (e.g. \textit{The Dan Bongino Show}).

\noindent\textbf{Distribution of the toxic chains across podcasts}: We compute the distribution of the toxic chains across the different podcast channels. Figure~\ref{fig:podcast_distribution} shows the distribution of top ten channels for both right- and left-leaning channels. The top \textit{five} contributing right-leaning podcast channels are \textit{Louder with Crowder}, \textit{Mark Levin Podcast}, \textit{The New Abnormal}, \textit{The Dan Bongino Show} and \textit{Get Off My Lawn Podcast w/ Gavin McInnes} in decreasing order of contribution, aggregating to a total of \textit{63.6\%}. From the left-leaning channels, \textit{Mea Culpa} and \textit{Lovett or Leave it} contribute nearly \textit{45.4\%} to toxic conversation chains. Together, these podcasts potentially reach tens of millions of listeners every week\footnote{e.g., \textit{Just Dan Bongino} and \textit{Mark Levin} podcasts are extremely popular with over 20 million weekly listeners (see \url{https://bit.ly/3ZEArcS})}. Many of these channels have been reported to be hosted and attended by people making homophobic, and racist remarks\footnote{\url{https://tinyurl.com/louder-crowder}}$^{,}$\footnote{\url{https://tinyurl.com/dan-bingo-show1}}$^{,}$\footnote{\url{https://tinyurl.com/dan-bingo-show2}}. 

% ------------------------------------------------------------

\section{Analysis of toxic conversation chains}
\label{section:analysis}

In this section we present an in-depth analysis of the segments constructed in the previous section.

\subsection{Basic properties}
We now study the basic textual properties of the diarized text corresponding to the anchor segments and compare them with the other segments in a chain. For all our textual analysis, we use the \texttt{PyNLPl}\footnote{\url{https://pypi.org/project/PyNLPl/}} library. In this section and Figure~\ref{fig:textual_statistics} we plot the results of right-leaning channels; similar analysis for left-leaning is presented in the Appendix.\\
\textbf{(i) Time coverage}: The duration of each segment in a conversation chain is the difference between the \texttt{start} and the \texttt{end} time obtained during diarization process. We calculate the mean of the segment's duration across all the chains and plot them in Figure~\ref{fig:textual_statistics}. Error bars indicate 95\% confidence intervals. Interestingly, the anchor segment has the highest mean implying that speakers tend to speak for longer duration when their speech is most toxic. Further, the rise in the mean duration is not sudden and neither is the fall. While the ascend starts from \texttt{Segment -1} itself, the decay is more gradual and flattens after \texttt{Segment 3}.\\
\textbf{(ii) Conversation organization}: Using the uni-gram probability values of the tokens we compute the entropy of the diarized text. Figure~\ref{fig:textual_statistics} shows that the anchor segment has a higher average entropy compared to the other segments indicating that the text in the anchor segment is more random and less well-formed. Further, we also calculate the perplexity of the diarized text assuming uni-gram probabilities.
Figure~\ref{fig:textual_statistics} shows that the mean perplexity is the highest for the anchor segment indicating that the conversation is least organized/coherent in this segment.

\begin{table}[!ht]
\setlength{\tabcolsep}{0.001mm}
\begin{tabular}{lll}
\textbf{\includegraphics[height=0.15\linewidth]{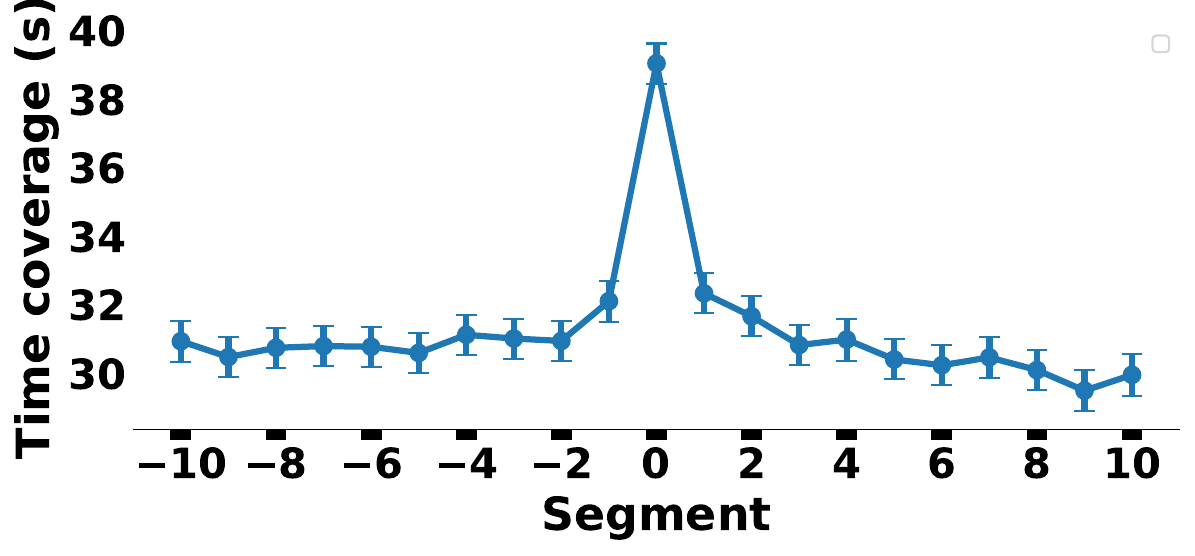}} &
\textbf{\includegraphics[height=0.15\linewidth]{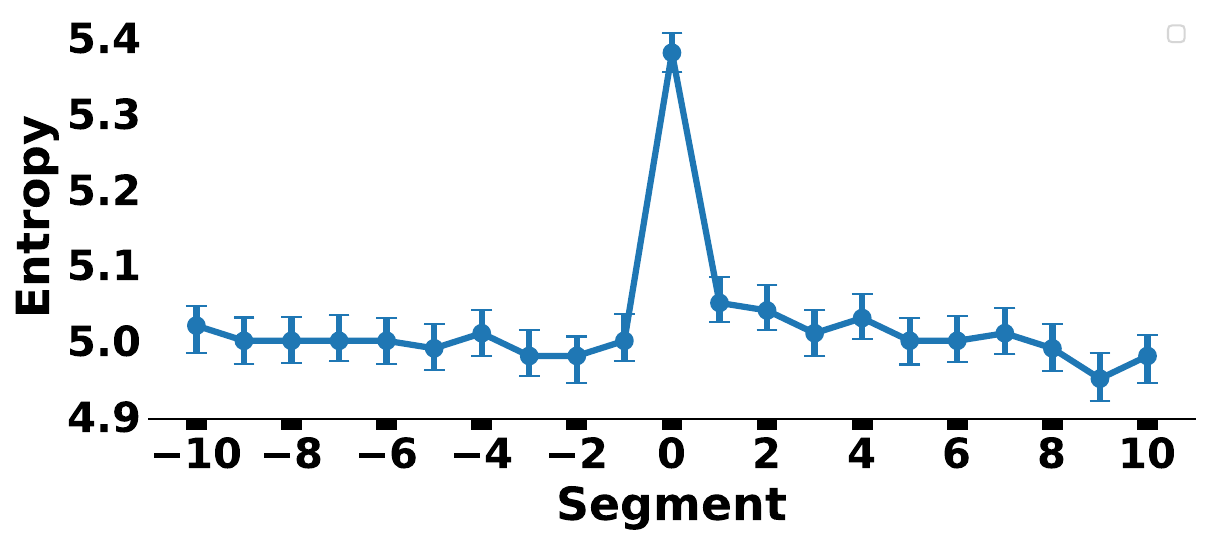}} & \textbf{\includegraphics[height=0.15\linewidth]{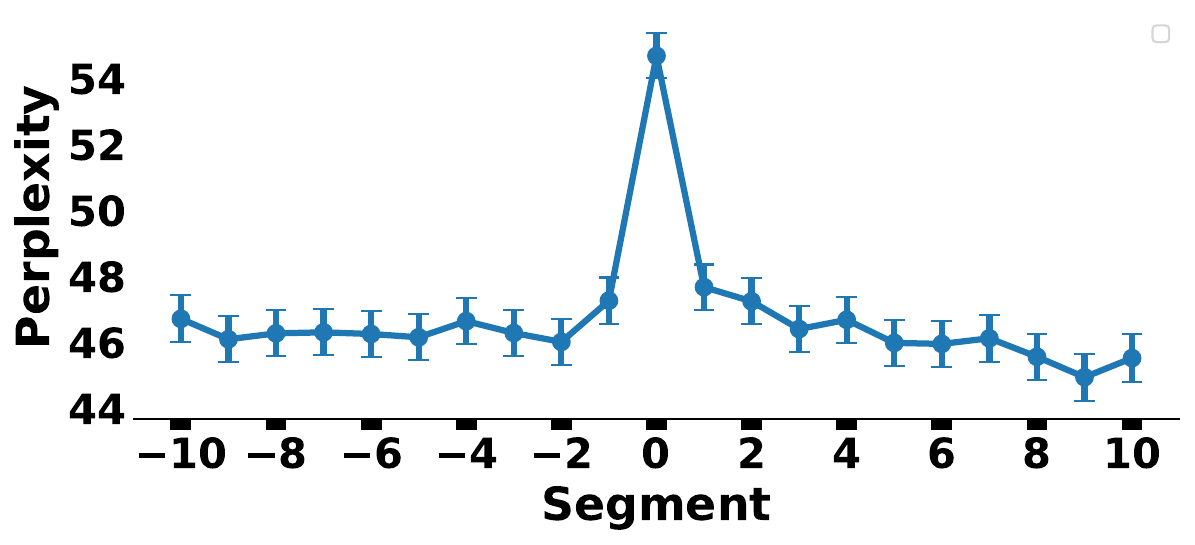}}
\end{tabular}
\captionof{figure}{\label{fig:textual_statistics}\footnotesize{\textsc{\textbf{Basic Properties:}} Mean at 95\% confidence intervals for time coverage, entropy and perplexity across segments.}}
\end{table}

\subsection{Figurative language}
Human conversation, especially in a public discourse like podcasts are often strewn with figurative language like hyperboles and metaphors~\cite{10.1111/comt.12096}. Hyperboles incorporate exaggeration for emphasis and metaphor is used to make implied comparison. In this section, we report the extent of such figurative language present in the conversation chains. We use the \texttt{bert-large-uncased} model finetuned on the \textsc{STL} task~\cite{badathala2023matchheavenmultitaskframework} to infer the hyperboles and metaphors from the diarized text. From Figure~\ref{fig:figurative_language}, we observe that, as expected, the usage of metaphors in podcasts is generally high ranging between 69.58\% -- 73.53\% for the non-anchor segments. Remarkably, it reaches to an all high of 81.91\% for the anchor segment. Further, we observe that the anchor segment has nearly double the percentage of hyperboles (18.03\%) compared to the highest among non-anchor segments (9.19\%).  Note that while these features are plotted for the right-leaning channels, the trends are very similar for the left-leaning ones. Thus, in summary, the toxic segment has the highest exaggeration level and more frequently invokes implied comparison. Both of these work as possible means to emphasize on the embedded toxic remark making it intense as well as subtle at the same time.

\begin{table}[!ht]
\setlength{\tabcolsep}{0.001mm}
\centering
\begin{tabular}{ll}
\textbf{\includegraphics[width=0.5\linewidth]{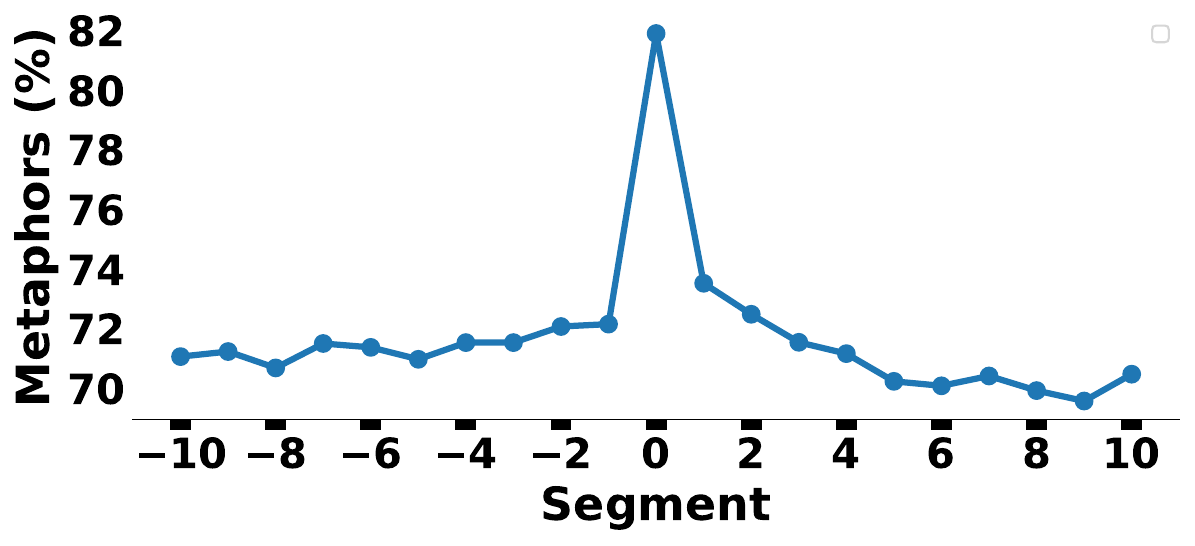}} & \textbf{\includegraphics[width=0.5\linewidth]{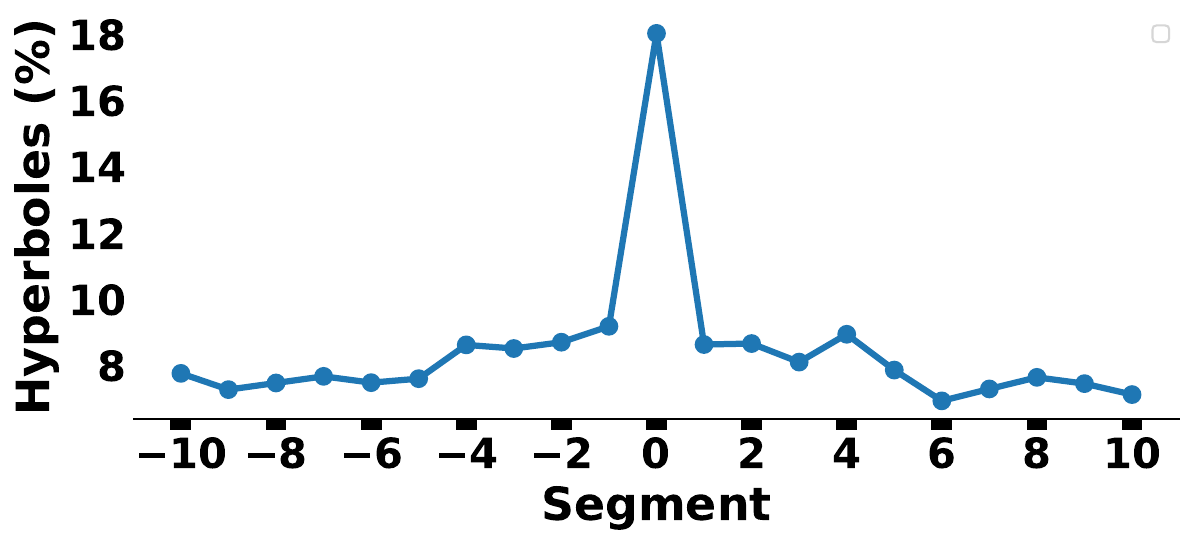}}
\end{tabular}
\captionof{figure}{\label{fig:figurative_language}\footnotesize{\textbf{\textsc{Figurative Language:}} Metaphors (left) and Hyperboles (right) across segments. The numbers are in percentage (\%).}}
\end{table}
\subsection{Empath features}
Empath is a metric which is frequently used in text analytics to assess the emotional content of the text. We also use empath as a metric to understand the underlying emotions and behavioral patterns latent in the conversation chains. Since our inputs are conversation chains and not flat text, we finetune the DistilBERT~\cite{sanh2020distilbertdistilledversionbert} model over the dataset presented in~\cite{rashkin2019empatheticopendomainconversationmodels} to infer the 32 empath features from these chains. In Figure~\ref{fig:empath_fig_1} we show the top eight most frequent emotions that have occurred at least 5\% of times within a segment. We observe considerably high intensity for the following empath features -- \textit{`angry', `furious', `annoyed', `disgusted'} in the anchor segment. On the other hand, there is a reduced intensity of the following empath features -- \textit{`surprised', `hopeful', `afraid', `anticipating'} in the anchor segment. Thus, toxic conversations are laden with aggression, resulting in the loss of decorum in the conversation. Note that while these features are plotted for the right-leaning channels, the trends are very similar for the left-leaning ones.

\setlength{\tabcolsep}{1.4mm}
\begin{table}[!ht]
\centering
\begin{tabular}{ll}
\textbf{\includegraphics[width=0.37\linewidth]{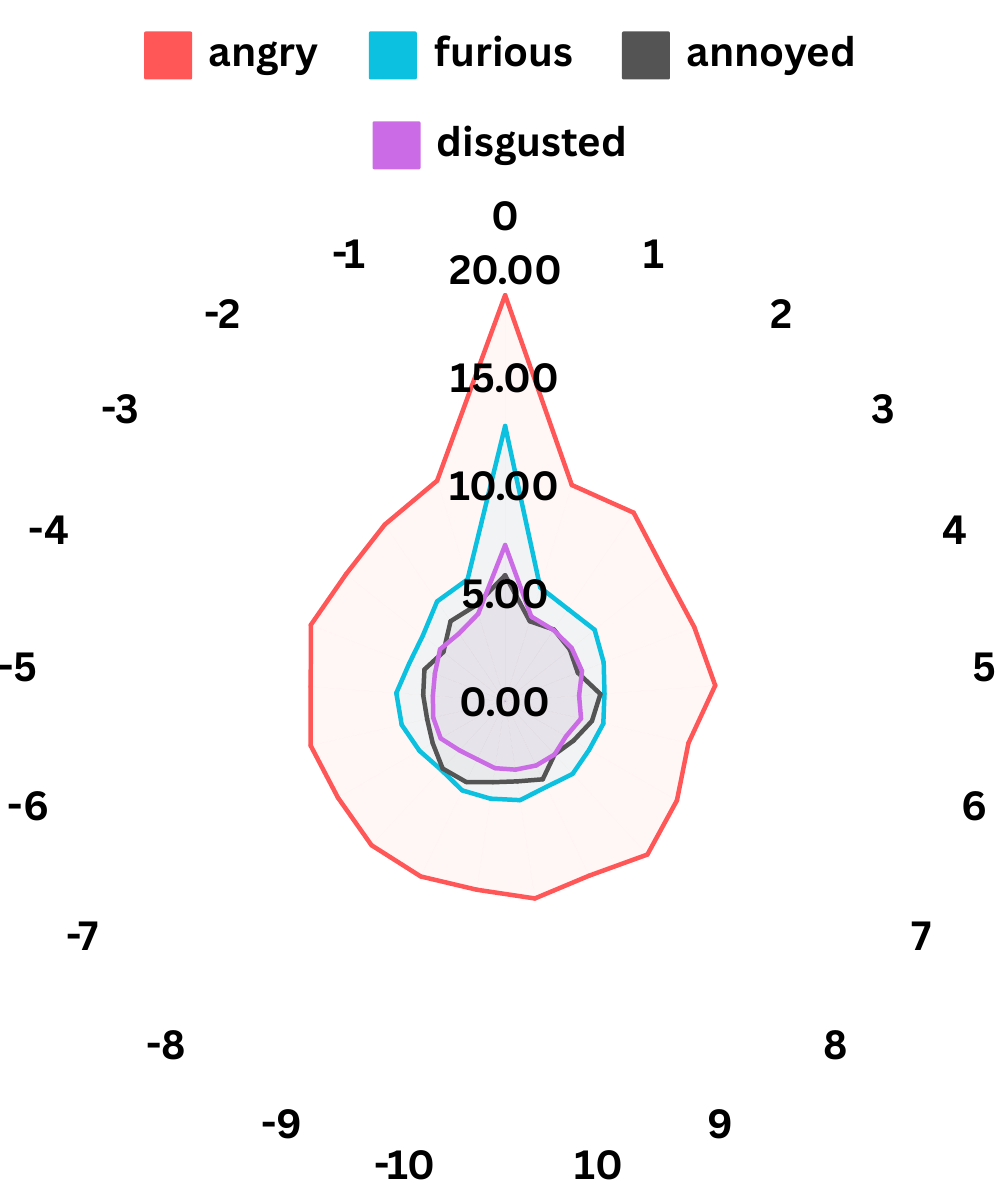}} & \textbf{\includegraphics[width=0.37\linewidth]{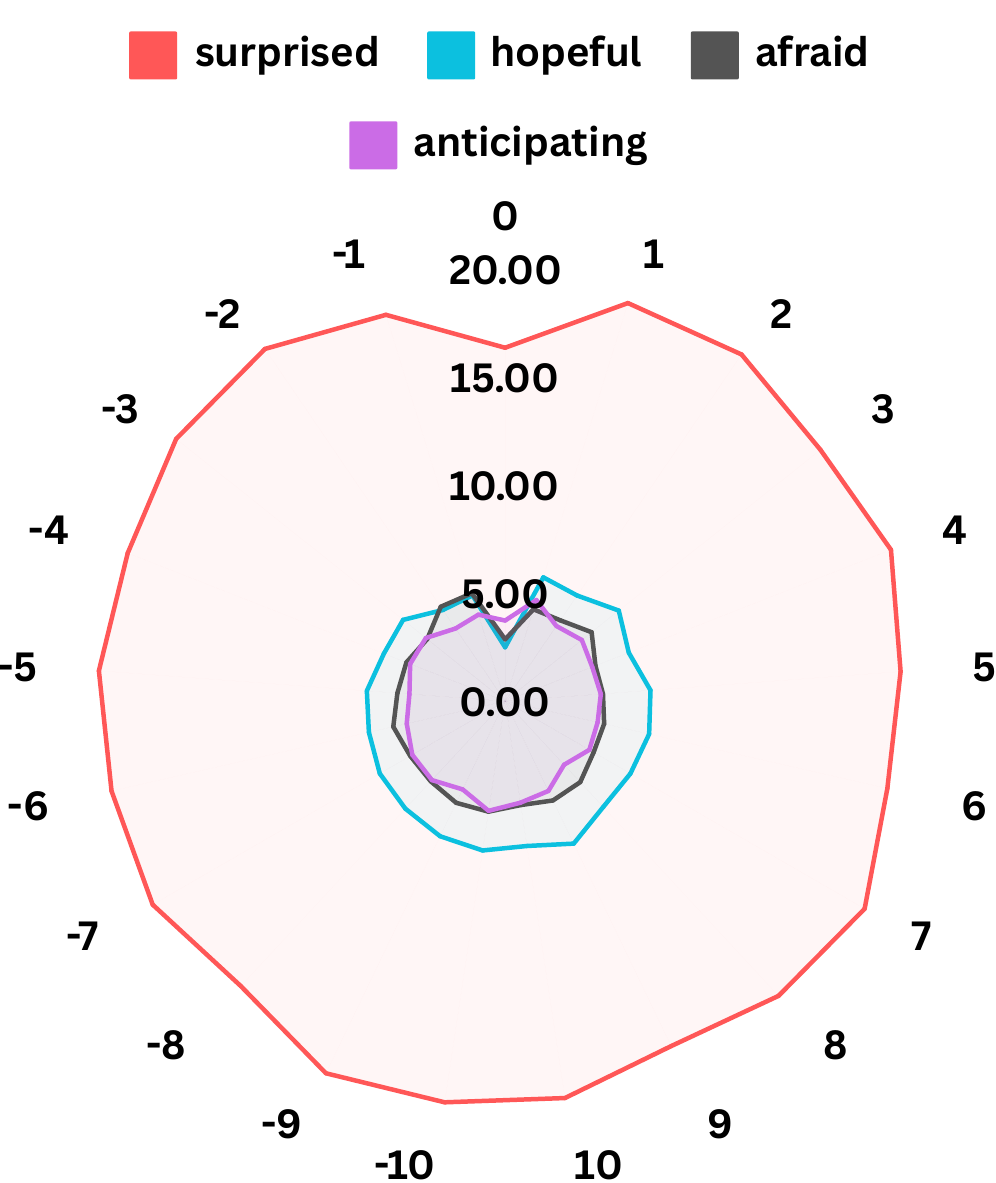}}
\end{tabular}
\captionof{figure}{\label{fig:empath_fig_1}\footnotesize{\textsc{\textbf{Empath Features:}} Plot on the left presents the features those which increased for anchor segment. Plot on the right illustrates the ones with decreased value. In each of these plots, anchor point is `0'; left \& right semicircles represent previous and next segments respectively.}}
\end{table}

\section{Toxicity begets toxicity: Change points in toxic chains}
\begin{table*}[!ht]
\centering
\scriptsize
\caption{\footnotesize{\textsc{\textbf{Cpd Results:}} Performance comparison of the traditional CPD algorithms, LLMs and audio-LLMs across different metrics and two different ways of aggregating the metrics (mean and median). Metrics for LLMs and audio-LLMs are averaged across three independent runs. Standard deviations (rounded to two decimal places) are \underline{\textit{underlined \& italicized}}; zero standard deviation is not put in the table to increase visibility. The numbers 1, 2 and 4 linked with precision and recall are the margins of errors allowed. The best results are highlighted in \colorbox[HTML]{CCF4CC}{\textbf{green}} for the right-leaning and in \colorbox[HTML]{ADD8E6}{\textbf{blue}} for the left-leaning channels. \textsc{\textbf{K-cpd}}: \textsc{KernelCPD}, \textsc{\textbf{B-up}}: \textsc{BottomUp}, \textsc{\textbf{B-seg}}: \textsc{BinSeg}, \textsc{\textbf{Qwen}}: \sysQWEN{}, \textsc{\textbf{Qwen-A}}: \sysQWENA{}, \textsc{\textbf{Gpt}}: \sysGPT{}, \textsc{\textbf{Gpt-A}}: \sysGPTA{}, \textbf{zs}: \textit{zero-shot}, \textbf{cot}: \textit{chain-of-thoughts}, L: left-leaning, \& R: right-leaning. $\downarrow$: Lower is better, $\uparrow$: Higher is better.}}
\begin{tabular}{cc|c|cccc|cccc|cccc|ccc}
\rowcolor[HTML]{D9EAD3} 
\multicolumn{2}{c|}{\cellcolor[HTML]{FFF2CC}\textbf{\textsc{Metrics}}}                                                                          & \textbf{\textsc{Agg}} & \textbf{\textsc{Pelt(R)}}                      & \textbf{\textsc{K-cpd(R)}}                        & \textbf{\textsc{B-up(R)}}                         & \textbf{\textsc{B-seg(R)}} & \textbf{\begin{tabular}[c]{@{}c@{}}\textsc{Qwen}\\ (zs-R)\end{tabular}} & \textbf{\begin{tabular}[c]{@{}c@{}}\textsc{Qwen}\\ (cot-R)\end{tabular}} & \textbf{\begin{tabular}[c]{@{}c@{}}\textsc{Qwen-a}\\ (zs-R)\end{tabular}} & \textbf{\begin{tabular}[c]{@{}c@{}}\textsc{Qwen-a}\\ (cot-R)\end{tabular}} & \textbf{\begin{tabular}[c]{@{}c@{}}GPT\\ (zs-R)\end{tabular}} & \textbf{\begin{tabular}[c]{@{}c@{}}GPT\\ (cot-R)\end{tabular}}                           & \textbf{\begin{tabular}[c]{@{}c@{}}GPT-A\\ (zs-R)\end{tabular}} & \textbf{\begin{tabular}[c]{@{}c@{}}GPT-A\\ (cot-R)\end{tabular}} & \textbf{\begin{tabular}[c]{@{}c@{}}\textsc{K-cpd}(L)\end{tabular}} & \textbf{\begin{tabular}[c]{@{}c@{}}GPT\\ (zs-L)\end{tabular}} & \textbf{\begin{tabular}[c]{@{}c@{}}GPT-A\\ (zs-L)\end{tabular}}\\ \hline
\multicolumn{2}{c|}{\cellcolor[HTML]{FFF2CC}}                                                                                          & \textbf{mean} & 7.03                               & 5.01                                  & 7.06                                  & 7.9            & \begin{tabular}[c]{@{}c@{}}5.4 \textit{\underline{.04}}\end{tabular}         & \begin{tabular}[c]{@{}c@{}}5.76 \textit{\underline{.04}}\end{tabular}         & 6.47                                                           & 7.96                                                            & \begin{tabular}[c]{@{}c@{}}4.16 \textit{\underline{.07}}\end{tabular}       & \cellcolor[HTML]{CCF4CC}\textbf{\begin{tabular}[c]{@{}c@{}}3.73 \textit{\underline{.03}}\end{tabular}} & \begin{tabular}[c]{@{}c@{}}3.88 \textit{\underline{.03}}\end{tabular}         & \begin{tabular}[c]{@{}c@{}}3.96 \textit{\underline{.02}}\end{tabular}   & 5.32    & \cellcolor[HTML]{ADD8E6}\textbf{3.98 \textit{\underline{.01}}} & 4.45 \textit{\underline{.03}}  \\
\multicolumn{2}{c|}{\multirow{-2}{*}{\cellcolor[HTML]{FFF2CC}\textbf{hausdorff} $\downarrow$}}                                                      & \textbf{med} & 7                                  & 4                                     & 7                                     & 8              & 5                                                            & 6                                                             & 7                                                              & 8                                                               & \cellcolor[HTML]{CCF4CC}\textbf{3}                          & \cellcolor[HTML]{CCF4CC}\textbf{3}                                                     & \cellcolor[HTML]{CCF4CC}\textbf{3}                            & \cellcolor[HTML]{CCF4CC}\textbf{3}                & 4.5         & \cellcolor[HTML]{ADD8E6}\textbf{4}  & \cellcolor[HTML]{ADD8E6}\textbf{4}  \\ \hline
\multicolumn{2}{c|}{\cellcolor[HTML]{FFF2CC}}                                                                                          & \textbf{mean} & 0.77                               & 0.84                                  & 0.74                                  & 0.57           & 0.78                                                         & 0.77                                                          & 0.78                                                           & 0.74                                                            & 0.86                                                        & \cellcolor[HTML]{CCF4CC}\textbf{0.87}                                                  & 0.86                                                          & 0.85             &         0.79    & \cellcolor[HTML]{ADD8E6}\textbf{0.86}                  & 0.85               \\
\multicolumn{2}{c|}{\multirow{-2}{*}{\cellcolor[HTML]{FFF2CC}\textbf{rand index} $\uparrow$}}                                                     & \textbf{med} & 0.78                               & 0.86                                  & 0.75                                  & 0.55           & 0.79                                                         & 0.78                                                          & 0.78                                                           & 0.75                                                            & 0.87                                                        & \cellcolor[HTML]{CCF4CC}\textbf{0.88}                                                  & \cellcolor[HTML]{CCF4CC}\textbf{0.88}                         & 0.87                    & 0.82                           & \cellcolor[HTML]{ADD8E6}\textbf{0.86}        & \cellcolor[HTML]{ADD8E6}\textbf{0.86}    \\ \hline
\multicolumn{1}{c|}{\cellcolor[HTML]{FFF2CC}}                                     & \cellcolor[HTML]{9698ED}                           & \textbf{mean} & 0.28                               & 0.47                                  & 0.24                                  & 0.14           & 0.23                                                         & 0.24                                                          & 0.28                                                           & 0.17                                                            & \begin{tabular}[c]{@{}c@{}}0.48 \textit{\underline{.01}}\end{tabular}       & 0.45                                                                                   & \cellcolor[HTML]{CCF4CC}\textbf{0.52}                         & 0.45                                        & 0.3         & 0.43     & \cellcolor[HTML]{ADD8E6}\textbf{0.47 \textit{\underline{.01}}}    \\
\multicolumn{1}{c|}{\multirow{-2}{*}{\cellcolor[HTML]{FFF2CC}\textbf{}}} & \multirow{-2}{*}{\cellcolor[HTML]{9698ED}1} & \textbf{med} & 0.27                               & \cellcolor[HTML]{CCF4CC}\textbf{0.5}  & 0.25                                  & 0              & \begin{tabular}[c]{@{}c@{}}0.26 \textit{\underline{.02}}\end{tabular}        & 0.29                                                          & 0.33                                                           & 0.17                                                            & \cellcolor[HTML]{CCF4CC}\textbf{0.5}                        & \begin{tabular}[c]{@{}c@{}}0.46 \textit{\underline{.06}}\end{tabular}                                  & \cellcolor[HTML]{CCF4CC}\textbf{0.5}                          & \cellcolor[HTML]{CCF4CC}\textbf{0.5}          & 0.25     & 0.36 \textit{\underline{.03}}      & \cellcolor[HTML]{ADD8E6}\textbf{0.5}      \\
\multicolumn{1}{c|}{\cellcolor[HTML]{FFF2CC}}                                     & \cellcolor[HTML]{9698ED}                           & \textbf{mean} & 0.35                               & 0.53                                  & 0.32                                  & 0.23           & 0.3                                                          & 0.32                                                          & 0.4                                                            & 0.31                                                            & 0.63                                                        & \begin{tabular}[c]{@{}c@{}}0.58 \textit{\underline{.01}}\end{tabular}                                  & \cellcolor[HTML]{CCF4CC}\textbf{0.66}                         & 0.6                      & 0.54                         & \cellcolor[HTML]{ADD8E6}\textbf{0.66}         & 0.65    \\
\multicolumn{1}{c|}{\multirow{-2}{*}{\cellcolor[HTML]{FFF2CC}\textbf{precision} $\uparrow$}} & \multirow{-2}{*}{\cellcolor[HTML]{9698ED}2} & \textbf{med} & 0.33                               & 0.5                                   & 0.25                                  & 0              & 0.29                                                         & 0.29                                                          & 0.33                                                           & 0.33                                                            & \cellcolor[HTML]{CCF4CC}\textbf{0.67}                       & \begin{tabular}[c]{@{}c@{}}0.6 \textit{\underline{.07}}\end{tabular}                                   & \cellcolor[HTML]{CCF4CC}\textbf{0.67}                         & 0.5                   & 0.5                           & \cellcolor[HTML]{ADD8E6}\textbf{0.67}        & \cellcolor[HTML]{ADD8E6}\textbf{0.67}      \\
\multicolumn{1}{c|}{\cellcolor[HTML]{FFF2CC}}                                     & \cellcolor[HTML]{9698ED}                           & \textbf{mean} & 0.42                               & 0.5                                   & 0.34                                  & 0.37           & 0.3                                                          & 0.35                                                          & 0.49                                                           & 0.38                                                            & \begin{tabular}[c]{@{}c@{}}0.61 \textit{\underline{.01}}\end{tabular}       & 0.56                                                                                   & \cellcolor[HTML]{CCF4CC}\textbf{0.67}                         & 0.6              & 0.57                       & 0.66              & \cellcolor[HTML]{ADD8E6}\textbf{0.72 \textit{\underline{.01}}}         \\
\multicolumn{1}{c|}{\multirow{-2}{*}{\cellcolor[HTML]{FFF2CC}\textbf{}}} & \multirow{-2}{*}{\cellcolor[HTML]{9698ED}4} & \textbf{med} & 0.33                               & 0.5                                   & 0.33                                  & 0.33           & 0.29                                                         & 0.29                                                          & 0.33                                                           & 0.4                                                             & \begin{tabular}[c]{@{}c@{}}0.52 \textit{\underline{.02}}\end{tabular}       & 0.5                                                                                    & \cellcolor[HTML]{CCF4CC}\textbf{0.67}                         & 0.5            & 0.5                            & \cellcolor[HTML]{ADD8E6}\textbf{0.67}        &    \cellcolor[HTML]{ADD8E6}\textbf{0.67}        \\ \hline
\multicolumn{1}{c|}{\cellcolor[HTML]{FFF2CC}}                                     & \cellcolor[HTML]{9698ED}                           & \textbf{mean} & 0.61                               & \cellcolor[HTML]{CCF4CC}\textbf{0.76} & 0.72                                  & 0.15           & 0.6                                                          & 0.61                                                          & 0.37                                                           & 0.37                                                            & \begin{tabular}[c]{@{}c@{}}0.51 \textit{\underline{.01}}\end{tabular}       & 0.54                                                                                   & 0.49                                                          & 0.5         & 0.45                           & \cellcolor[HTML]{ADD8E6}\textbf{0.48}             &     0.42      \\
\multicolumn{1}{c|}{\multirow{-2}{*}{\cellcolor[HTML]{FFF2CC}\textbf{}}}    & \multirow{-2}{*}{\cellcolor[HTML]{9698ED}1} & \textbf{med} & 0.6                                & \cellcolor[HTML]{CCF4CC}\textbf{1}    & 0.75                                  & 0              & 0.5                                                          & 0.5                                                           & 0.33                                                           & 0.33                                                            & 0.5                                                         & 0.5                                                                                    & 0.5                                                           & 0.5         & \cellcolor[HTML]{ADD8E6}\textbf{0.5}               & \cellcolor[HTML]{ADD8E6}\textbf{0.5}        & 0.4                            \\
\multicolumn{1}{c|}{\cellcolor[HTML]{FFF2CC}}                                     & \cellcolor[HTML]{9698ED}                           & \textbf{mean} & 0.75                               & 0.85                                  & \cellcolor[HTML]{CCF4CC}\textbf{0.93} & 0.24           & 0.8                                                          & 0.8                                                           & 0.53                                                           & 0.63                                                            & \begin{tabular}[c]{@{}c@{}}0.69 \textit{\underline{.01}}\end{tabular}       & \begin{tabular}[c]{@{}c@{}}0.7 \textit{\underline{.01}}\end{tabular}                                   & 0.63                                                          & 0.65            & \cellcolor[HTML]{ADD8E6}\textbf{0.71}                  & 0.69 \textit{\underline{.01}}                & 0.56             \\
\multicolumn{1}{c|}{\multirow{-2}{*}{\cellcolor[HTML]{FFF2CC}\textbf{recall} $\uparrow$}}    & \multirow{-2}{*}{\cellcolor[HTML]{9698ED}2} & \textbf{med} & 0.75                               & \cellcolor[HTML]{CCF4CC}\textbf{1}    & \cellcolor[HTML]{CCF4CC}\textbf{1}    & 0              & \cellcolor[HTML]{CCF4CC}\textbf{1}                           & \cellcolor[HTML]{CCF4CC}\textbf{1}                            & 0.5                                                            & 0.67                                                            & 0.67                                                        & 0.67                                                                                   & \begin{tabular}[c]{@{}c@{}}0.53 \textit{\underline{.02}}\end{tabular}         & 0.6                  & \cellcolor[HTML]{ADD8E6}\textbf{0.75}           & 0.67        & 0.5                       \\
\multicolumn{1}{c|}{\cellcolor[HTML]{FFF2CC}}                                     & \cellcolor[HTML]{9698ED}                           & \textbf{mean} & 0.89                               & 0.82                                  & \cellcolor[HTML]{CCF4CC}\textbf{0.98} & 0.34           & 0.8                                                          & 0.88                                                          & 0.66                                                           & 0.79                                                            & 0.69                                                        & 0.69                                                                                   & 0.66                                                          & 0.66                & \cellcolor[HTML]{ADD8E6}\textbf{0.74}                     & 0.69 \textit{\underline{.01}}          & 0.61            \\
\multicolumn{1}{c|}{\multirow{-2}{*}{\cellcolor[HTML]{FFF2CC}\textbf{}}}    & \multirow{-2}{*}{\cellcolor[HTML]{9698ED}4} & \textbf{med} & \cellcolor[HTML]{CCF4CC}\textbf{1} & \cellcolor[HTML]{CCF4CC}\textbf{1}    & \cellcolor[HTML]{CCF4CC}\textbf{1}    & 0.33           & 0.8                                                          & \cellcolor[HTML]{CCF4CC}\textbf{1}                            & 0.55                                                           & 0.78                                                            & \begin{tabular}[c]{@{}c@{}}0.64 \textit{\underline{.02}}\end{tabular}       & \begin{tabular}[c]{@{}c@{}}0.66 \textit{\underline{.02}}\end{tabular}                                  & \begin{tabular}[c]{@{}c@{}}0.61 \textit{\underline{.02}}\end{tabular}         & \begin{tabular}[c]{@{}c@{}}0.57 \textit{\underline{.02}}\end{tabular} & \cellcolor[HTML]{ADD8E6}\textbf{0.75} & 0.67       & 0.5
\end{tabular}
\label{tab:cpd-results}
\end{table*}

\begin{table}[!ht]
\centering
\begin{tabular}{l}
\textbf{\includegraphics[height=0.28\linewidth]{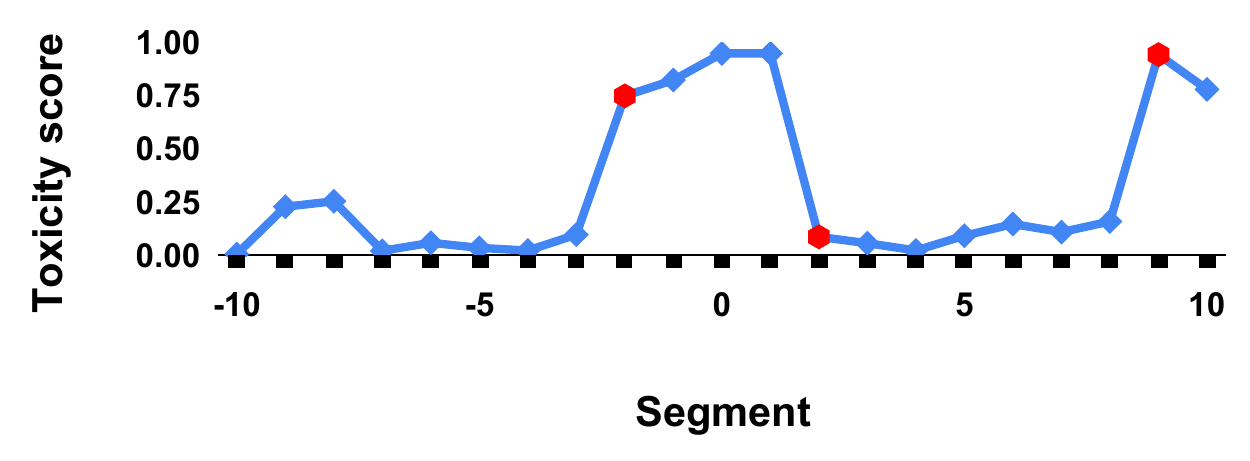}} \\
\textbf{\includegraphics[height=0.28\linewidth]{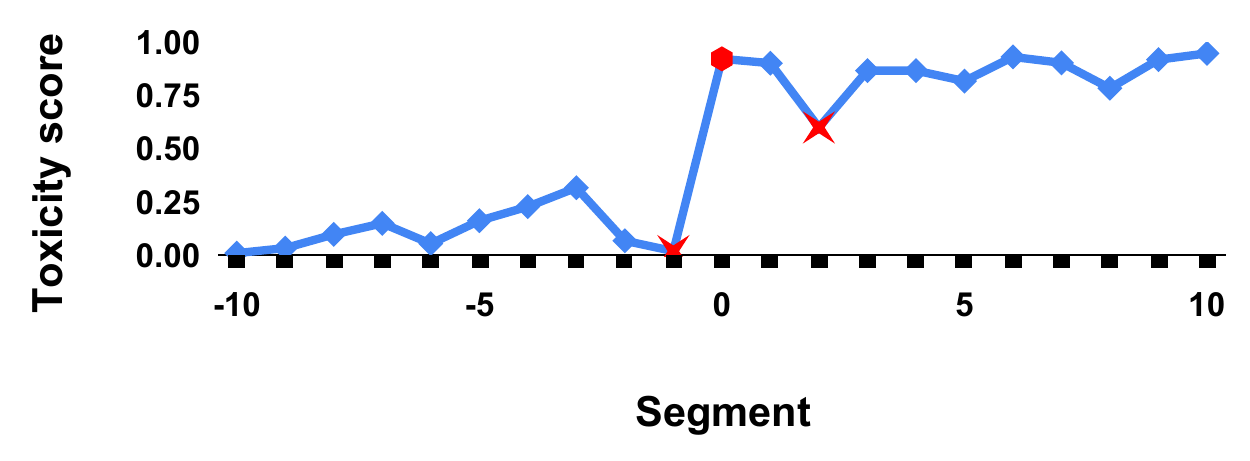}}
\end{tabular}
\captionof{figure}{\textsc{\textbf{Cpd Examples:}}\label{fig:cpd_examples} \footnotesize{Plots representing two samples comparing human annotation with \sysGPTA{} \textit{zero-shot} setup's detected change points. Correctly predicted points are marked with a \textcolor{red}{red} hexagon and incorrect predictions are marked with \textcolor{red}{red} cross.}}
\end{table}

\label{section:cpd}
While linguistic properties studied in the previous section enables us to characterize the conversation chains, it is not possible to formulate actionable insights for designing effective interventions. To achieve this, one has to acknowledge that toxic conversation chains are inherently dynamic, with evolving interactions often involving multiple participants. Understanding the trajectory of these conversations requires identifying key points where significant shifts occur. Automatic change point detection can enable researchers to: \textbf{\textit{(i) Isolate key moments}}: Automatically identify segments of conversations where toxicity sharply increases/decreases or new topics are introduced; \textbf{\textit{(ii) Understand contextual triggers}}: Track down the patterns latent in events or statements that precede toxicity spikes or topic changes and thus reveal actionable insights for designing intervention strategies.
\subsection{Automatic change point detection}
First, we employ standard change point detection (CPD) algorithms to automatically identify change points within conversation chains. Specifically, we consider various search methods -- \textsc{Pelt}, \textsc{KernelCPD}, \textsc{Window}, \textsc{BottomUp} \& \textsc{Binseg} alongside different cost functions including \texttt{rbf}, \texttt{cosine}, \texttt{l2} \& \texttt{linear}. All implementations are done through \texttt{ruptures}~\cite{TRUONG2020107299} library.\footnote{\url{https://centre-borelli.github.io/ruptures-docs/}} 
Among all the possible combinations over these search methods and cost functions; only \texttt{rbf} as the cost function coupled with \textsc{Pelt}, \textsc{KernelCPD}, \textsc{BottomUp} \& \textsc{Binseg} search methods successfully identifies at least one change point. As a result, we conduct our experiments to identify change points using these four algorithms. Limitations of primitive CPD algorithms are their inability to process a conversation by taking into account diarized text and audio as well. Hence as a next step, we extend our experiments to LLMs and audio-LLMs in two different setups: \textit{zero-shot} and \textit{chain-of-thoughts}. Details of the employed models are provided below and the complete experimental setup is provided in Section~\ref{section:reproducibility}. Due to paucity of space, the employed prompts are provided in the Appendix.\\
\textbf{Employed LLMs \& audio-LLMs}: We use \sysQWEN{}~\cite{qwen2}\footnote{\url{https://huggingface.co/Qwen/Qwen2-7B-Instruct}} \& \sysGPT{} for running textual prompts where we pass the diarized text and toxicity score (refer to Appendix for detailed prompts) of each segment as input to the models. \sysQWENA{}~\cite{chu2024qwen2} \& \sysGPTA{} can process audio inputs (\textit{.wav} files) as well along with textual prompt for conditional generation. These models use an audio encoder in combination with an LLM for conditional generation. While \sysQWENA{} uses Whisper large-v3 model~\cite{radford2022whisper} as audio encoder and \textsc{Qwen-7B}~\cite{qwen} as the LLM backbone; architectural details of \sysGPTA{} are unknown since it is a proprietary model. We employ the following checkpoints for these models -- \sysQWEN{}: \texttt{Qwen/Qwen2-7B-Instruct}, \sysQWENA{}:\footnote{\texttt{Qwen/Qwen2-Audio-7B-Instruct}} \& \sysGPTA{}:\footnote{\texttt{gpt-4o-audio-preview}}. Although we have taken proper care of reproducibility (refer Section~\ref{section:reproducibility}), to demonstrate the robustness of the approach we run each model and prompt setup for three independent runs. In Table~\ref{tab:cpd-results} we present the mean and standard deviation of the metrics for LLMs and audio-LLMs.

\subsection{Manual annotation of change points} 
To evaluate the predictive performance of the baseline CPD algorithms we conduct a manual annotation exercise of the change points. We consider as many as top 100 toxic conversation chains based on the toxicity scores of the anchor segments from each of the right- and left-leaning channels. We then get all the change points in each of these chains annotated by three annotators. All three annotators are experts in hate speech analysis research.\\
\textbf{Annotation instructions:} To ensure consistency and reliability in the annotation process, we ask the annotators to base their judgments on the following factors -- \textbf{(a)} \textit{tone}: shifts in sentiment or intensity, such as transitions from neutral or mildly toxic statements to overt hostility, \textbf{(b)} \textit{topical shift}: the emergence of new discussion themes or the cessation of previously dominant topics and \textbf{(c)} \textit{change in toxicity}: escalations or reductions in the degree of harmful language or expressions. Further since this is a sensitive task we instruct the annotators to \textbf{(a)} maintain confidentiality and handle the conversational data responsibly, \textbf{(b)} focus on objective evaluation without imposing personal biases and \textbf{(c)} annotate at most 15 data points per day to ensure annotation quality and mental well being of annotators. We appropriately remunerate the annotators with Amazon gift vouchers.\\
Overall, the three annotators combined marked 79\% \& 80\% of instances to contain 2, 3 or 4 change points for right- and left-leaning channels, respectively, with the minimum and maximum being 1 and 7, respectively. The total number of change points marked by them across all the 100 chains is 982 for right- and 923 for left-leaning channels. We believe that this is a unique and a first-of-its-kind dataset that can be used in future to train/evaluate other predictive algorithms.
\subsection{Experimental setup and reproducibility}
\label{section:reproducibility}
Temperature controls the randomness in the generation of LLMs with \textit{near-zero} value meaning less randomness across independent runs, hence leading to high reproducibility. Therefore, we run all our experiments with a very low temperature value of 0.001. To further ensure reproducibility, we run the LLMs and audio-LLMs three time with the exact same prompts. The results in the Table~\ref{tab:cpd-results} are presented with mean and standard deviation for these models. For most cases, the standard deviation is near zero demonstrating the robustness of the system. Coming to implementation details, \sysQWEN{} and \sysQWENA{} are coded using \textsc{PyTorch} library in \textsc{Python}. \sysGPT{} and \sysGPTA{} are proprietary models and we use them through the provided APIs. Throughout our experiments, we set \textit{max\_new\_tokens} parameter to 50 across all prompt variants tried on all LLMs and audio-LLMs. We use \sysQWEN{} and \sysQWENA{} models through HuggingFace APIs after accepting their license agreements and for \sysQWENA{} we also apply 8-bit quantization using bitsandbytes\footnote{\url{https://huggingface.co/docs/bitsandbytes/en/index}} library due to resource constraints. For both of the open source models we utilize Kaggle's 2\textsc{x} T4 GPU servers, each having GPU memory of 15GB. These GPUs have a usage limit of 30GB for 30 hours per week.

\subsection{Performance evaluation}
\textbf{Evaluation metrics}: In order to measure the correspondence between the manually annotated change points and those generated by the CPD algorithms, LLMs \& audio-LLMs we use four standard metrics -- \textit{Hausdorff distance},\footnote{\url{https://en.wikipedia.org/wiki/Hausdorff_distance}} \textit{rand index}, \textit{precision}, and \textit{recall}. While calculating the metrics, if a particular change point is marked by a majority of the annotators, it is considered a valid change point. While the total number of unique points annotated by all three annotators is 614 (458), after the majority voting we arrive at 266 (309) points, for right (left)-leaning channels respectively. The predictions of the algorithms are evaluated against these majority voted change points for evaluation. \\
\noindent\textbf{Key results (right-leaning)}: \sysGPTA{} with zero-shot prompting for a margin of 4 is the best model among the considered CPD algorithms and LLMs/audio-LLMs. It has the highest precision across all margins and has much better \textit{Hausdorff distance} and \textit{rand index} than the best CPD algorithm -- \textsc{KernelCPD}. Surprisingly, in the case of both \sysGPT{} and \sysGPTA{}, there is little to no improvement in precision with \textit{chain-of-thoughts}. \sysQWEN{} and \sysQWENA{} under-performs when compared to the \textsc{KernelCPD} algorithm. Interestingly, \sysGPT{} and \sysGPTA{} do not generate very high recall values compared to other experimental setups, which is natural as these models only generate relevant change points, compared to other methods/models. This further enhances their robustness \& usability for such a complex task.

\noindent Two representative examples of the change points across two different chains are provided in Figure~\ref{fig:cpd_examples}.

\noindent\textbf{Key results (left-leaning)}:  Table~\ref{tab:cpd-results} sheds light on change point detection results on the top 100 most toxic chosen from the left-leaning channels. Since \textsc{Kernel-cpd} is the best traditional method and the zero-shot prompting on \sysGPT{} and \sysGPTA{} perform effectively on right-leaning channels, we repeat only those setups here as well to save computational costs. Aligning with the previous results, \sysGPTA{} outperforms \sysGPT{} and \textsc{Kernel-CPD} in terms of precision with a margin of 4 and has competing scores with \sysGPT{} for other metrics. These results not only demonstrate the robustness of generative AI models but also ground their impressive real-time content moderation capabilities for this immensely important medium.

% ------------------------------------------------------------

\begin{table}[!ht]
\centering
\setlength{\tabcolsep}{1.4mm}
\scriptsize
\caption{\footnotesize{\textsc{\textbf{Alternate Measures}}: Performance comparison over different toxicity scoring models for \textsc{KernelCPD} and \textit{zero-shot} variants of \sysGPT{} and \sysGPTA{}. Results of \textsc{Perspective-API} are as picked from Table~\ref{tab:cpd-results}. The numbers 1, 2 and 4 linked with precision and recall are the margins of errors allowed. The best results are \colorbox[HTML]{CCF4CC}{\textbf{highlighted}}. \textsc{\textbf{K-C}}: \textsc{KernelCPD}, \textsc{\textbf{G}}: \sysGPT{}, \textsc{\textbf{G-A}}: \sysGPTA{}, \textsc{AGG}: Aggregate, \textsc{P-A}: Perspective API, \textsc{T-o}: \texttt{toxic-bert-original} and \textsc{T-u}: \texttt{toxic-bert-unbiased}. $\downarrow$: Lower is better, $\uparrow$: Higher is better.}}
\begin{tabular}{cc|c|ccc|cccccc}
\rowcolor[HTML]{D9EAD3} 
\multicolumn{2}{c|}{\cellcolor[HTML]{FFF2CC}}                                                                                            & \cellcolor[HTML]{D9EAD3}                               & \textbf{K-C}                        & \textbf{K-C}                       & \textbf{K-C}                        & \textbf{G}                          & \textbf{G}                          & \textbf{G}                          & \textbf{G-A}                        & \textbf{G-A}                        & \textbf{G-A}                        \\
\rowcolor[HTML]{D9EAD3} 
\multicolumn{2}{c|}{\multirow{-2}{*}{\cellcolor[HTML]{FFF2CC}\textbf{\textsc{Metrics}}}}                                                          & \multirow{-2}{*}{\cellcolor[HTML]{D9EAD3}\textbf{AGG}} & \textbf{\textsc{P-A}}                        & \textbf{\textsc{T-o}}                      & \textbf{\textsc{T-u}}                        & \textbf{\textsc{P-A}}                        & \textbf{\textsc{T-o}}                       & \textbf{\textsc{T-u}}                        & \textbf{\textsc{P-A}}                        & \textbf{\textsc{T-o}}                       & \textbf{\textsc{T-u}}                        \\ \hline
\multicolumn{2}{c|}{\cellcolor[HTML]{FFF2CC}}                                                                                            & \textbf{mean}                                          & 5.01                                  & 4.3                                  & 5.22                                  & 4.16                                  & \cellcolor[HTML]{CCF4CC}\textbf{3.79} & 4.02                                  & 3.88                                  & 4                                     & 4.25                                  \\
\multicolumn{2}{c|}{\multirow{-2}{*}{\cellcolor[HTML]{FFF2CC}\textbf{hausdorff} $\downarrow$}}                                                        & \textbf{med}                                           & 4                                     & 4                                    & 4                                     & \cellcolor[HTML]{CCF4CC}\textbf{3}    & \cellcolor[HTML]{CCF4CC}\textbf{3}    & \cellcolor[HTML]{CCF4CC}\textbf{3}    & \cellcolor[HTML]{CCF4CC}\textbf{3}    & 4                                     & 4                                     \\ \hline
\multicolumn{2}{c|}{\cellcolor[HTML]{FFF2CC}}                                                                                            & \textbf{mean}                                          & 0.84                                  & 0.85                                 & 0.82 & \cellcolor[HTML]{CCF4CC}\textbf{0.86}                                  & \cellcolor[HTML]{CCF4CC}\textbf{0.86}                                  & \cellcolor[HTML]{CCF4CC}\textbf{0.86}                                  & \cellcolor[HTML]{CCF4CC}\textbf{0.86}                                  & 0.84                                  & 0.84                                  \\
\multicolumn{2}{c|}{\multirow{-2}{*}{\cellcolor[HTML]{FFF2CC}\textbf{rand index} $\uparrow$}}                                                       & \textbf{med}                                           & 0.86                                  & 0.87                                 & 0.84 & 0.87                                  & \cellcolor[HTML]{CCF4CC}\textbf{0.88}                                  & 0.87                                  & \cellcolor[HTML]{CCF4CC}\textbf{0.88}                                  & 0.87                                  & 0.87                                  \\ \hline
\multicolumn{1}{c|}{\cellcolor[HTML]{FFF2CC}}                                     & \cellcolor[HTML]{9698ED}                             & \textbf{mean}                                          & 0.47                                  & 0.5                                  & 0.42                                  & 0.48                                  & 0.48                                  & 0.49                                  & \cellcolor[HTML]{CCF4CC}\textbf{0.52} & 0.47                                  & 0.49                                  \\
\multicolumn{1}{c|}{\cellcolor[HTML]{FFF2CC}}                                     & \multirow{-2}{*}{\cellcolor[HTML]{9698ED}\textbf{1}} & \textbf{med}                                           & \cellcolor[HTML]{CCF4CC}\textbf{0.5}  & \cellcolor[HTML]{CCF4CC}\textbf{0.5} & 0.43                                  & \cellcolor[HTML]{CCF4CC}\textbf{0.5}  & \cellcolor[HTML]{CCF4CC}\textbf{0.5}  & \cellcolor[HTML]{CCF4CC}\textbf{0.5}  & \cellcolor[HTML]{CCF4CC}\textbf{0.5}  & \cellcolor[HTML]{CCF4CC}\textbf{0.5}  & \cellcolor[HTML]{CCF4CC}\textbf{0.5}  \\
\multicolumn{1}{c|}{\cellcolor[HTML]{FFF2CC}}                                     & \cellcolor[HTML]{9698ED}                             & \textbf{mean}                                          & 0.53                                  & 0.58                                 & 0.49                                  & 0.63                                  & 0.64                                  & 0.6                                   & \cellcolor[HTML]{CCF4CC}\textbf{0.66} & 0.64                                  & 0.6                                   \\
\multicolumn{1}{c|}{\cellcolor[HTML]{FFF2CC}}                                     & \multirow{-2}{*}{\cellcolor[HTML]{9698ED}\textbf{2}} & \textbf{med}                                           & 0.5                                   & 0.5                                  & 0.5                                   & \cellcolor[HTML]{CCF4CC}\textbf{0.67} & \cellcolor[HTML]{CCF4CC}\textbf{0.67} & 0.55                                  & \cellcolor[HTML]{CCF4CC}\textbf{0.67} & 0.6                                   & 0.5                                   \\
\multicolumn{1}{c|}{\cellcolor[HTML]{FFF2CC}}                                     & \cellcolor[HTML]{9698ED}                             & \textbf{mean}                                          & 0.5                                   & 0.58                                 & 0.5                                   & 0.61                                  & 0.64                                  & 0.63                                  & \cellcolor[HTML]{CCF4CC}\textbf{0.67} & \cellcolor[HTML]{CCF4CC}\textbf{0.67} & 0.66                                  \\
\multicolumn{1}{c|}{\multirow{-6}{*}{\cellcolor[HTML]{FFF2CC}\textbf{precision} $\uparrow$}} & \multirow{-2}{*}{\cellcolor[HTML]{9698ED}\textbf{4}} & \textbf{med}                                           & 0.5                                   & 0.5                                  & 0.5                                   & 0.52                                  & \cellcolor[HTML]{CCF4CC}\textbf{0.67} & \cellcolor[HTML]{CCF4CC}\textbf{0.67} & \cellcolor[HTML]{CCF4CC}\textbf{0.67} & \cellcolor[HTML]{CCF4CC}\textbf{0.67} & \cellcolor[HTML]{CCF4CC}\textbf{0.67} \\ \hline
\multicolumn{1}{c|}{\cellcolor[HTML]{FFF2CC}}                                     & \cellcolor[HTML]{9698ED}                             & \textbf{mean}                                          & \cellcolor[HTML]{CCF4CC}\textbf{0.76} & 0.67                                 & 0.72                                  & 0.51                                  & 0.56                                  & 0.58                                  & 0.49                                  & 0.49                                  & 0.53                                  \\
\multicolumn{1}{c|}{\cellcolor[HTML]{FFF2CC}}                                     & \multirow{-2}{*}{\cellcolor[HTML]{9698ED}\textbf{1}} & \textbf{med}                                           & \cellcolor[HTML]{CCF4CC}\textbf{1}    & 0.67                                 & 0.75                                  & 0.5                                   & 0.5                                   & 0.5                                   & 0.5                                   & 0.5                                   & 0.5                                   \\
\multicolumn{1}{c|}{\cellcolor[HTML]{FFF2CC}}                                     & \cellcolor[HTML]{9698ED}                             & \textbf{mean}                                          & \cellcolor[HTML]{CCF4CC}\textbf{0.85} & 0.77                                 & 0.81                                  & 0.69                                  & 0.74                                  & 0.7                                   & 0.63                                  & 0.66                                  & 0.65                                  \\
\multicolumn{1}{c|}{\cellcolor[HTML]{FFF2CC}}                                     & \multirow{-2}{*}{\cellcolor[HTML]{9698ED}\textbf{2}} & \textbf{med}                                           & \cellcolor[HTML]{CCF4CC}\textbf{1}    & \cellcolor[HTML]{CCF4CC}\textbf{1}   & \cellcolor[HTML]{CCF4CC}\textbf{1}    & 0.67                                  & 0.75                                  & 0.67                                  & 0.53                                  & 0.67                                  & 0.67                                  \\
\multicolumn{1}{c|}{\cellcolor[HTML]{FFF2CC}}                                     & \cellcolor[HTML]{9698ED}                             & \textbf{mean}                                          & 0.82                                  & 0.78                                 & \cellcolor[HTML]{CCF4CC}\textbf{0.83} & 0.69                                  & 0.73                                  & 0.74                                  & 0.66                                  & 0.69                                  & 0.71                                  \\
\multicolumn{1}{c|}{\multirow{-6}{*}{\cellcolor[HTML]{FFF2CC}\textbf{recall} $\uparrow$}}    & \multirow{-2}{*}{\cellcolor[HTML]{9698ED}\textbf{4}} & \textbf{med}                                           & \cellcolor[HTML]{CCF4CC}\textbf{1}    & 0.8                                  & \cellcolor[HTML]{CCF4CC}\textbf{1}    & 0.64                                  & 0.71                                  & 0.71                                  & 0.61                                  & 0.67                                  & 0.67                                 
\end{tabular}
\label{tab:ablation}
\end{table}
% ------------------------------------------------------------

\subsection{CPD based on alternate toxicity scores}
We performed extensive change point detection experiments based on the toxicity scores from Perspective API in the previous subsections. Here, we repeat those experiments with the scores obtained from \textsc{Detoxify} library on the best-performing methods, i.e., \textsc{Kernel-CPD}, \sysGPT{} (zero-shot) \& \sysGPTA{} (zero-shot) on right-leaning most toxic conversation chains (the results for the left-leaning channels are very similar). To obtain the toxicity scores, we use the two \textsc{Toxic-BERT} variants -- \texttt{original} and \texttt{unbiased} as noted in Section~\ref{sec:toxic_conversation_chains}. Table~\ref{tab:ablation} outlines the detailed results from these experiments. Some of the key observations are as follows.\\
\textbf{(i)} The toxicity scores obtained from different tools have very little impact on the predictions made by \sysGPT{} and \sysGPTA{}. This is because the two models also take as input the source diarized text (\sysGPT{}) or the source audio snippet (\sysGPTA{}) each of which serves as additional context for the CPD.\\
\textbf{(ii)} The observation that \sysGPTA{} is the best model when toxicity scores are taken from Perspective API (margin = 4) is unequivocally corroborated when the other two measures of toxicity are taken (from \texttt{toxic-bert-original} and \texttt{toxic-bert-biased}). This further strengthens Perspective API's usability as a toxicity score generator.\\

% ------------------------------------------------------------

\section{Concluding discussion}
Rigorous research projects on toxicity mitigation have lead to the development of various strategies like counter-speech~\cite{Mathew_Saha_Tharad_Rajgaria_Singhania_Maity_Goyal_Mukherjee_2019}, text detoxification~\cite{dementieva2024overview}, and meme intervention~\cite{jha2024memeguardllmvlmbasedframework}, among many others~\cite{rizwan-etal-2025-hateprism}, including some controversial strategies~\cite{leerssen2023end}. Podcasts are comparatively different, since the interaction of the audience with guests/hosts is nearly negligible. Also, unlike social media content, they are generally characterized by long conversations and figuring out exact toxic segments in real-time is difficult. Thus, for current scenarios, intervention is limited to only beeping or stripping away such contents which are not real-time and are often ignored by podcast owners and streaming platforms due to the lack of automation strategies. Further, unlike access to social media, podcasts can easily be made available in an offline mode and can create a much bigger indirect impact. In addition, the toxic clips can be spread on social media without any further context thus spicing up the whole thing and leading to even worse polarization.

\noindent In this paper we introduced a dataset of transcribed political podcasts and formulated toxic conversation chains from the transcribed text. First, we make the surprising observation on the high prevalence of toxicity \textit{(detected with high confidence)} in podcasts which reach tens of millions of users. Next, we made various interesting observations regarding the linguistic characteristics of these chains. We noted that the anchor segment corresponding to the most toxic part of the chain has a lot of repetition, is less well-formed and has emotions and sentiments related to anger and fury. Expressions of demand in conversations can quickly lead the discourse to higher levels of toxicity. Subsequently, we manually annotate 200 toxic chains with change points for training and evaluation of automatic methods. Based on this ground-truth dataset we established that \sysGPT{}, \sysGPTA{} or CPD algorithms like \textsc{KernelCPD} can be effectively used to monitor the toxicity levels of conversation chains in real-time. The choice of one over the other can be made based on the budget available and the desired precision.\\
\noindent The CPD experiments as demonstrated here can be used to automatically monitor the toxicity levels in real-time. Such real-time monitoring can be very useful in controlling toxicity by making the participating hosts/guests aware of the rising toxicity in their conversation through a recommendation/alert system. Based on adherence to such recommendations, points, badges and other incentives may be awarded to the participants to promote healthy interaction and deescalate toxicity.  In cases of extreme violation, automatic termination of the recording of the conversation can also be implemented.

\section{Acknowledgments}
We thank Subhankar Swain, an MS (by research) student at the Indian Institute of Technology Kharagpur, for helping us with the CPD annotations. Kiran Garimella would like to acknowledge help from Ishan Mishra for the data collection.

% ------------------------------------------------------------

\bibliographystyle{ACM-Reference-Format}
\balance
\bibliography{references}

\appendix

\section{Limitations}

As future work, we plan to extend our study to incorporate audio signals as well in our analysis including a thorough investigation of the different audio features. We also plan to develop intervention strategies for this complex scenario.
Further we plan to expand our dataset and scalable tooling to other political podcasts to get clear estimates on the prevalence of toxicity in podcasts.
Finally, an important aspect that our paper does not answer is the impact of such toxicity on the listeners. Our findings lead us to question whether toxicity might become normalised if coming from popular podcast hosts who reach tens of millions of people, and the consequences it might have on our political discourse. Social scientists can study and answer such questions using some of the tools we developed in this work. .

% ------------------------------------------------------------

\section{Ethics statement}
Our dataset comprises popular podcasts curated from publicly available RSS feeds, which are widely accessible and listened to by tens of millions of people. These podcasts are analyzed using automated transcription and diarization techniques. While individual speakers are segmented to enable conversational analysis, their identities are neither inferred nor used in the analysis to ensure privacy. We recognize the potential for biases in various stages of our pipeline, including transcription, diarization, toxicity detection, and change point detection. To mitigate these biases, we employed a rigorous evaluation process, comparing multiple algorithms at each stage and selecting the best-performing ones based on empirical results. While our analysis yielded promising findings, we emphasize the importance of conducting a comprehensive audit of the pipeline to further evaluate its reliability and fairness, particularly in production or high-stakes contexts. Our study adheres to ethical guidelines for research on publicly available data and prioritizes transparency, privacy, and the minimization of harm. We aim to contribute constructively to the discourse on toxicity in podcasts while acknowledging the limitations of our methods and the need for ongoing scrutiny and improvement.

% ------------------------------------------------------------

\begin{table*}[!ht]
\footnotesize
\centering
\setlength{\tabcolsep}{1.5mm}
\renewcommand{\arraystretch}{1.2}
\caption{\footnotesize{\label{tab:dataset_stats}\textbf{\textsc{Complete Dataset Statistics:}} We crawl a total of 52 podcast channels (31 right- and 21 left-leaning), making a total of 12,322 episodes (with 9,166 from right- and 3,156 from left-leaning channels)}. Statistics for the crawled dataset sorted according to the number of episodes are presented. Columns for duration (in minutes) and mean token count are shown. In these columns, numbers in parenthesis specify the standard deviation. Columns for the number of toxic episodes (those containing at least one toxic conversation chain) and the corresponding percentage distribution are also provided. First 31 channels are right-leaning followed by 21 left-leaning channels.}
\begin{tabular}{
>{\columncolor[HTML]{CBCEFB}}l |ccccc}
\multicolumn{1}{c|}{\cellcolor[HTML]{D9D9D9}\textbf{Podcast channel name}} &
  \multicolumn{1}{c|}{\cellcolor[HTML]{D9D9D9}\textbf{\begin{tabular}[c]{@{}c@{}}Number\\ of episodes\end{tabular}}} &
  \multicolumn{1}{c|}{\cellcolor[HTML]{D9D9D9}\textbf{\begin{tabular}[c]{@{}c@{}}Average episode\\ duration (min)\end{tabular}}} &
  \multicolumn{1}{c|}{\cellcolor[HTML]{D9D9D9}\textbf{\begin{tabular}[c]{@{}c@{}}Average words\\ count\end{tabular}}} &
  \multicolumn{1}{c|}{\cellcolor[HTML]{D9D9D9}\textbf{\begin{tabular}[c]{@{}c@{}}Toxic\\ episodes\end{tabular}}} &
  \cellcolor[HTML]{D9D9D9}\textbf{\begin{tabular}[c]{@{}c@{}}\% toxic\\ episodes\end{tabular}} \\ \hline
\textbf{Bannon s War Room}                         & 1184 & 54 \textit{(2)}   & 9339 \textit{(662)}   & 172 & 15  \\
\textbf{Bill O Reilly s No Spin News and Analysis} & 1117 & 18 \textit{(17)}  & 2553 \textit{(2460)}  & 72  & 6   \\
\textbf{The Sean Hannity Show}                     & 990  & 37 \textit{(5)}  & 6375 \textit{(983)}  & 208 & 21  \\
\textbf{The Glenn Beck Program}                    & 725  & 81 \textit{(37)}  & 12619 \textit{(5703)} & 275 & 38  \\
\textbf{The Dan Bongino Show}                      & 448  & 48 \textit{(15)}  & 8483 \textit{(2706)}  & 309 & 69  \\
\textbf{Mark Levin Podcast}                        & 440  & 103 \textit{(18)} & 14280 \textit{(3095)} & 380 & 86  \\
\textbf{Human Events Daily with Jack Posobiec}     & 403  & 27 \textit{(8)}  & 4614 \textit{(1602)}  & 34  & 8   \\
\textbf{Conservative Review with Daniel Horowitz}  & 398  & 62 \textit{(6)}  & 9673 \textit{(1210)}  & 95  & 24  \\
\textbf{The Rubin Report}                          & 372  & 43 \textit{(13)}  & 8280 \textit{(2513)}  & 161 & 43  \\
\textbf{The News Why It Matters}                   & 334  & 44 \textit{(0)}  & 8313 \textit{(423)}  & 181 & 54  \\
\textbf{The Megyn Kelly Show}                      & 332  & 95 \textit{(10)}  & 17626 \textit{(2117)} & 180 & 54  \\
\textbf{Tim Pool Daily Show}                       & 321  & 86 \textit{(11)}  & 15472 \textit{(2295)} & 114 & 36  \\
\textbf{The Ben Shapiro Show}                      & 229  & 48 \textit{(17)}  & 10133 \textit{(3664)} & 62  & 27  \\
\textbf{The New Abnormal}                          & 220  & 50 \textit{(16)}  & 9132 \textit{(2919)}  & 198 & 90  \\
\textbf{Louder with Crowder}                       & 210  & 62 \textit{(33)}  & 12588 \textit{(6900)} & 186 & 89  \\
\textbf{The Charlie Kirk Show}                     & 200  & 40 \textit{(15)}  & 7127 \textit{(2903)}  & 30  & 15  \\
\textbf{The Michael Savage Show}                   & 196  & 54 \textit{(19)}  & 9253 \textit{(3384)}  & 96  & 49  \\
\textbf{The Jordan B Peterson Podcast}             & 145  & 99 \textit{(21)}  & 16236 \textit{(3772)} & 27  & 19  \\
\textbf{The Michael Knowles Show}                  & 133  & 46 \textit{(17)}  & 8112 \textit{(3176)}  & 43  & 32  \\
\textbf{Verdict with Ted Cruz}                     & 133  & 43 \textit{(11)}  & 7142 \textit{(1833)}  & 25  & 19  \\
\textbf{Hold These Truths with Dan Crenshaw}       & 115  & 53 \textit{(20)}  & 9699 \textit{(3478)}  & 13  & 11  \\
\textbf{Bret Weinstein DarkHorse Podcast}          & 94   & 106 \textit{(21)} & 17414 \textit{(3496)} & 39  & 41  \\
\textbf{Pseudo Intellectual with Lauren Chen}      & 88   & 11 \textit{(2)}  & 2202 \textit{(379)}  & 10  & 11  \\
\textbf{Fireside Chat with Dennis Prager}          & 87   & 32 \textit{(7)}  & 4309 \textit{(1168)}  & 13  & 15  \\
\textbf{Conversations With Coleman}                & 82   & 71 \textit{(19)}  & 12920 \textit{(3681)} & 22  & 27  \\
\textbf{The One w Greg Gutfeld}                    & 71   & 15 \textit{(3)}  & 2690 \textit{(645)}  & 34  & 48  \\
\textbf{Candace Owens}                             & 47   & 29 \textit{(14)}  & 5620 \textit{(2782)}  & 13  & 28  \\
\textbf{Get Off My Lawn Podcast w Gavin McInnes}   & 24   & 61 \textit{(20)}  & 10012 \textit{(3391)} & 24  & 100 \\
\textbf{The MeidasTouch Podcast}                   & 15   & 52 \textit{(40)}  & 9017 \textit{(6961)}  & 9   & 60  \\
\textbf{The Matt Walsh Show}                       & 10   & 46 \textit{(24)}  & 8081 \textit{(4228)}  & 3   & 30  \\
\textbf{Rudy Giuliani s Common Sense}              & 3    & 36 \textit{(4)}  & 5414 \textit{(750)}  & 2   & 67  \\ \hline \hline
\textbf{The MediasTouch Podcast}                   & 431  & 29 \textit{(27)}  & 4883 \textit{(4748)}  & 115 & 27  \\
\textbf{Late Night with Seth Meyers Podcast}       & 255  & 25 \textit{(5)}  & 4616 \textit{(840)}  & 73  & 29  \\
\textbf{Mea Culpa}                                 & 231  & 78 \textit{(12)}  & 12858 \textit{(2126)} & 225 & 97  \\
\textbf{Pod Save America}                          & 230  & 60 \textit{(16)}  & 11499 \textit{(2972)} & 187 & 81  \\
\textbf{In the Bubble with Andy Slavitt}           & 224  & 48 \textit{(10)}  & 8300 \textit{(1843)}  & 14  & 6   \\
\textbf{Fast Politics with Molly Jong-Fast}        & 201  & 55 \textit{(9)}  & 10036 \textit{(1617)} & 141 & 70  \\
\textbf{The Rachel Maddow Show}                    & 141  & 55 \textit{(28)}  & 9496 \textit{(4538)}  & 23  & 16  \\
\textbf{On with Kara Swisher}                      & 135  & 53 \textit{(10)}  & 10302 \textit{(2077)} & 64  & 47  \\
\textbf{Political Gabfest}                         & 132  & 51 \textit{(13)}  & 9041 \textit{(2330)}  & 17  & 13  \\
\textbf{Pod Save the World}                        & 124  & 67 \textit{(16)}  & 12273 \textit{(2935)} & 77  & 62  \\
\textbf{Lovett or Leave It}                        & 119  & 70 \textit{(21)}  & 12423 \textit{(3618)} & 118 & 99  \\
\textbf{Why is This Happening with Chris Hayes}    & 116  & 53 \textit{(13)}  & 9933 \textit{(2348)}  & 7   & 6   \\
\textbf{Majority 54}                               & 110  & 52 \textit{(10)}  & 10087 \textit{(2025)} & 35  & 32  \\
\textbf{Krystal Kyle and Friends}                  & 108  & 80 \textit{(17)}  & 14409 \textit{(3101)} & 104 & 96  \\
\textbf{Hysteria}                                  & 105  & 71 \textit{(15)}  & 12658 \textit{(2629)} & 100 & 95  \\
\textbf{Offline with Jon Favreau}                  & 97   & 54 \textit{(13)}  & 10041 \textit{(2733)} & 39  & 40  \\
\textbf{Conversations With Coleman}                & 96   & 67 \textit{(20)}  & 12524 \textit{(3805)} & 23  & 24  \\
\textbf{Pod Save the People}                       & 92   & 68 \textit{(17)}  & 11598 \textit{(3039)} & 12  & 13  \\
\textbf{Hell and High Water with John Heilemann}   & 72   & 72 \textit{(13)}  & 14766 \textit{(2853)} & 57  & 79  \\
\textbf{Intercepted with Jeremy Scahill}           & 70   & 48 \textit{(14)}  & 7672 \textit{(2537)}  & 4   & 6   \\
\textbf{Lady Dont Take No}                         & 67   & 45 \textit{(8)} & 7828 \textit{(1645)}  & 63  & 94 
\end{tabular}
\end{table*}

% ------------------------------------------------------------

\section{Employed prompts}

\begin{figure*}[!ht]
    \centering
    \includegraphics[width=1.85\columnwidth]{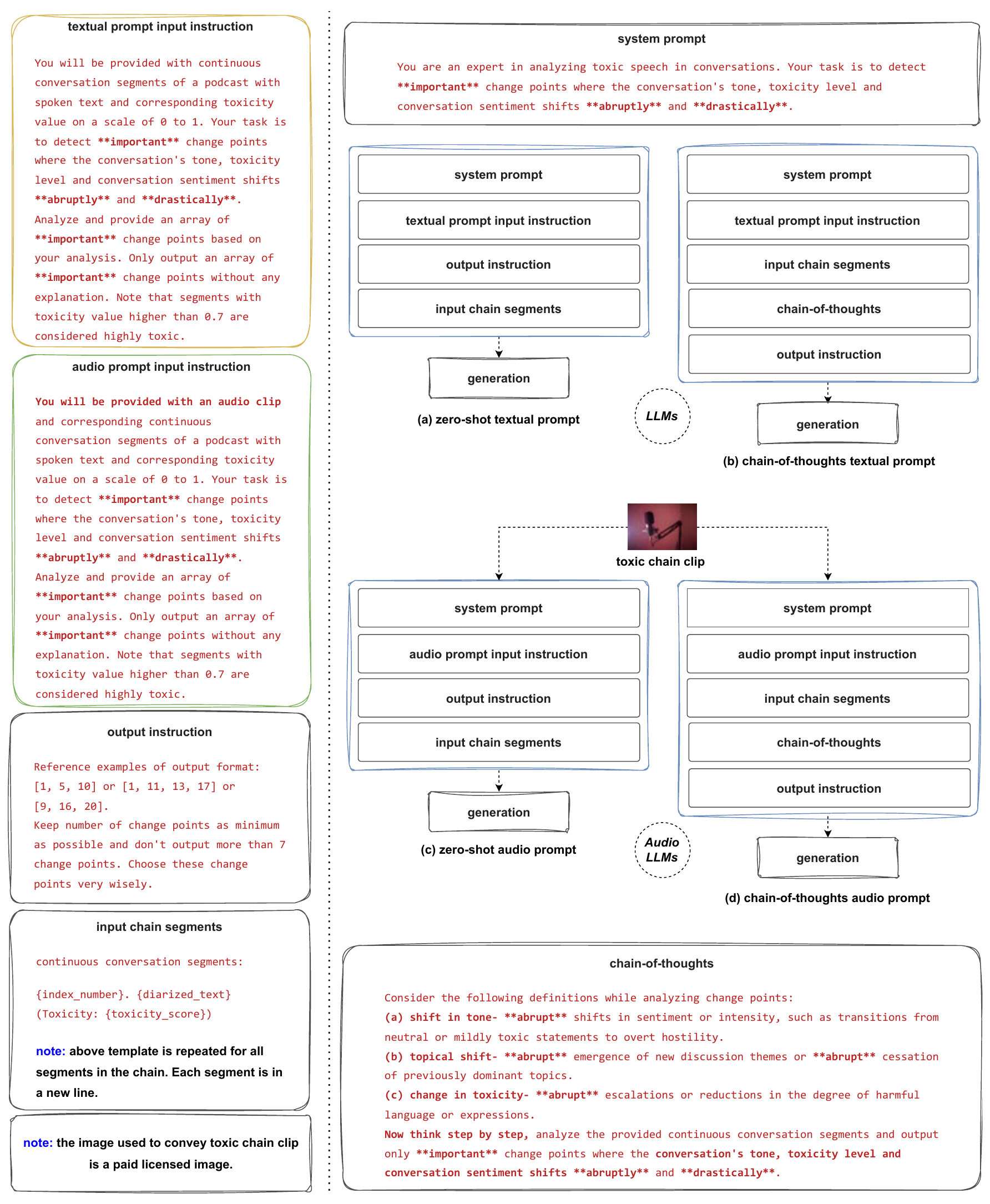}
    \caption{\footnotesize{\textbf{\textsc{Employed Prompts:}} Textual prompts used on \sysQWEN{} and \sysGPT{} are in sub-figures (a) and (b). Similarly, sub-figures (c) and (d) are employed with \sysQWENA{} and \sysGPTA{}. Different components of an input prompt are separated within individual boxes. Each of the sub-figures (i.e. (a), (b), (c) \& (d)) provides an overview of the organization of prompt components in different input prompt setups, i.e., in the \textit{zero-shot} \& \textit{chain-of-thoughts} setups. Please refer to Figure~\ref{fig:conversation_examples_2} and a figure in main paper for examples of input chain segments.}}
    \label{fig:prompts_employed}
\end{figure*}

\label{sec:prompts}

\textbf{Toxicity classifier prompt on \sysGPT{}}: We apply the following prompt on anchor segment's diarized text:\\
\noindent\texttt{\textbf{- ``role"}: ``system", \textbf{``content"}: ``You are an expert in toxic speech detection. Your task is to detect whether a provided speech is toxic or not."}\\
\noindent\texttt{\textbf{- ``role"}: ``user", \textbf{``type"}: ``text", \textbf{``content"}: ``Definition of toxicity: Perspective API’s primary attribute is toxicity, defined as a rude, disrespectful, or unreasonable comment that is likely to make you leave a discussion.\\
You are provided with the following speech: \{anchor segment's diarized text\}\\
Based on the above definition, classify the speech as either toxic or not-toxic. Output only toxic or not-toxic based on your analysis."}\\
\noindent\textbf{Change point detection prompts}: An overview of the employed prompts is presented in Figure~\ref{fig:prompts_employed}. The figure illustrates the combination of different prompt components gathered together to run textual \& audio-based LLMs. In the case of audio, toxic chain clip (\textit{.wav} file format) across all segments combined is also provided as input.

% ------------------------------------------------------------

\begin{table*}[!ht]
\centering
\begin{tabular}{ll}
\textbf{\includegraphics[width=1.0\linewidth]{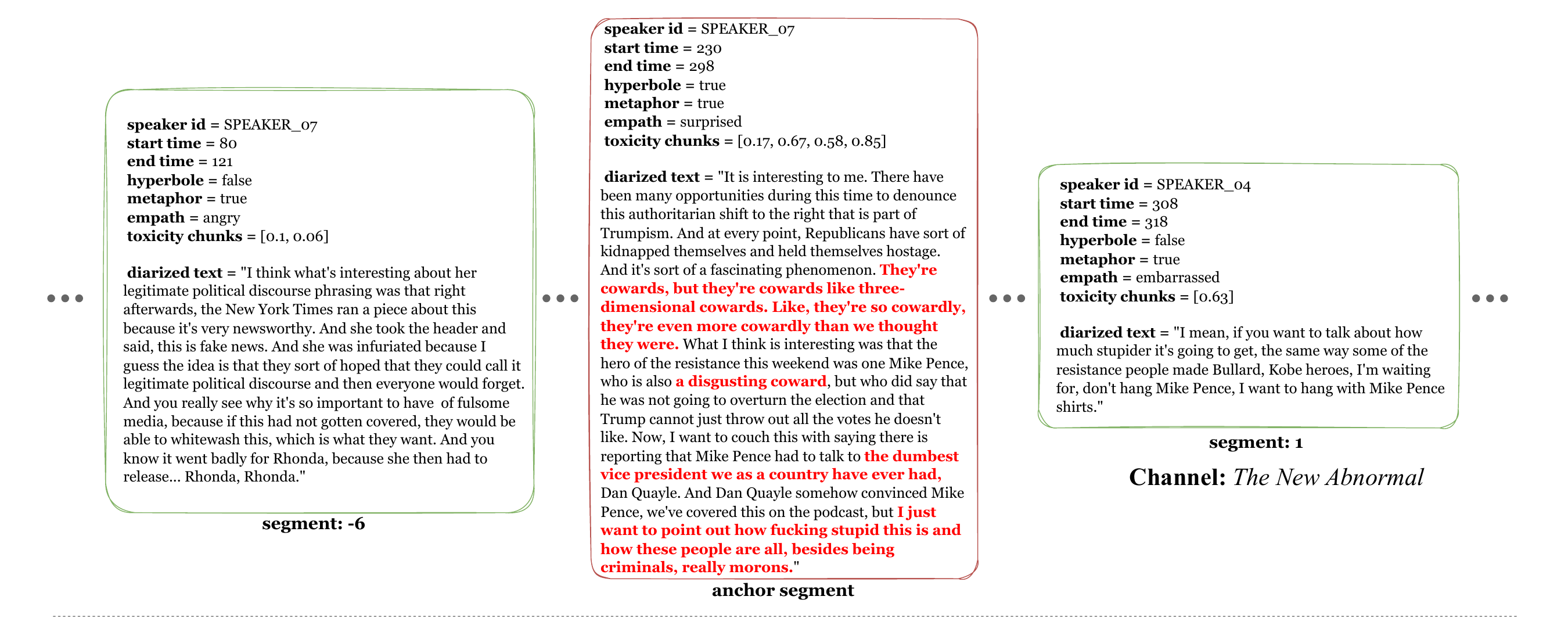}} \\
\textbf{\includegraphics[width=1.0\linewidth]{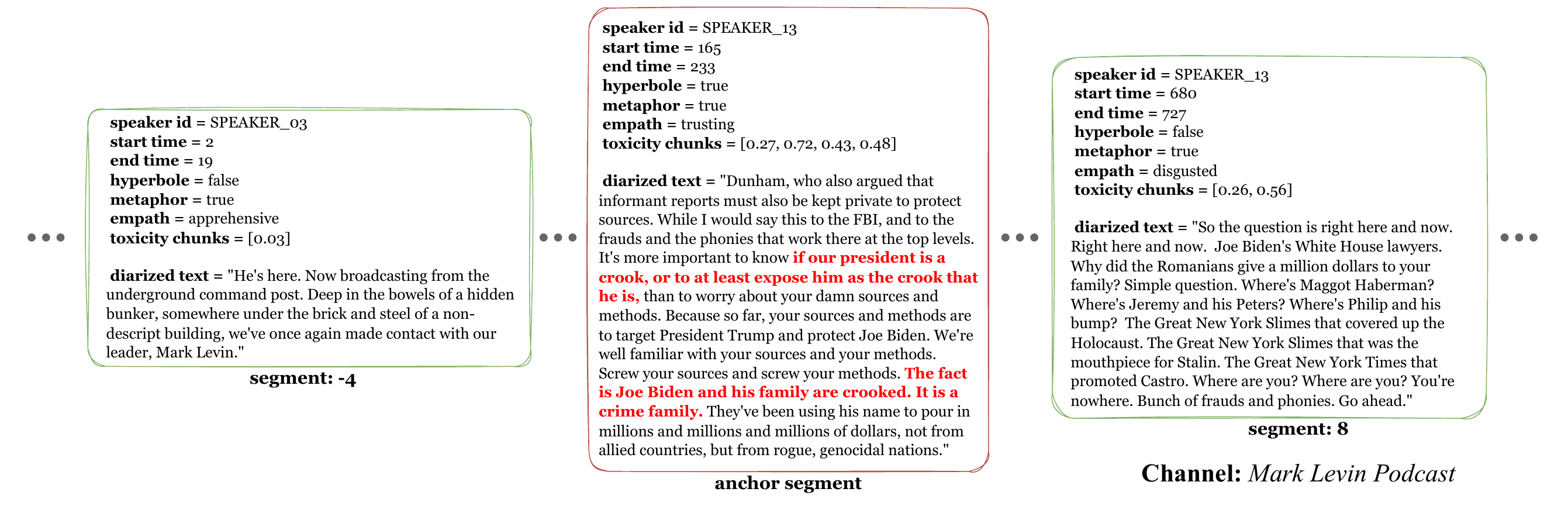}}
\end{tabular}
\captionof{figure}{\label{fig:conversation_examples_2} \footnotesize{\textbf{\textsc{Conversation Examples}}: One among previous and next segments are plotted along with anchor segment since it is not feasible to plot all segments. Toxic contents in the anchor segment are marked in \textcolor{red}{red} color. \textsc{Note:} start and end times are in seconds.}}
\end{table*}

% ------------------------------------------------------------

%\section{\nr{Analysis: left-leaning channels and right-leaning control group}}
\section{Properties of left-leaning channels}
The statistics of the 7,124 toxic conversation chains from left-leaning podcast channels are presented in Figure~\ref{fig:left_statistics}. Interestingly, we observe that these properties highly align with the findings we had previously covered for right-leaning podcast channels (in the main text). Thus the dynamics of toxicity appears to be universal across the two leanings. %Hence we robustly tunnel down that the inherent characteristics of toxic conversation chains are almost similar, irrespective of the right- or left- leaning behavior of the conversation. Further, it also points to a common human reaction in handling such instances.

\section{Further analysis for right-leaning channels}
\label{sec:further_analysis}
In this section we present some additional analysis of the right-leaning channels (the results are very similar for the left-leaning ones and hence not shown).
\subsection{Additional textual properties}
\textbf{(i) Token count}: We calculate the total number of tokens in the diarized text for each segment in a conversation chain. To break the words into tokens, we use the \textit{word\_tokenize} utility from NLTK library. Figure~\ref{fig:more_textual_statistics} shows that the mean token count is the largest for the anchor segment which naturally follows from the earlier observation that these segments have the longest mean duration.\\
\noindent\textbf{(ii) Type token ratio (TTR)}: This metric is used to evaluate the diversity of vocabulary in a text. It is given by the formula: $\frac{\textrm{\#types}} {\textrm{\#tokens}}$, where $types$ are unique $tokens$. Higher values indicate greater lexical diversity which means that the text has less redundancy. Lower values suggest repetitive textual content. Figure~\ref{fig:more_textual_statistics} shows that the mean TTR for anchor segment is lower than all other segments in the chain, indicating that there is more repetition (possibly of the same hateful remark) in the anchor segment.

\setlength{\tabcolsep}{0.001mm}
\begin{table}[!ht]
\begin{tabular}{ll}
\textbf{\includegraphics[width=0.5\linewidth]{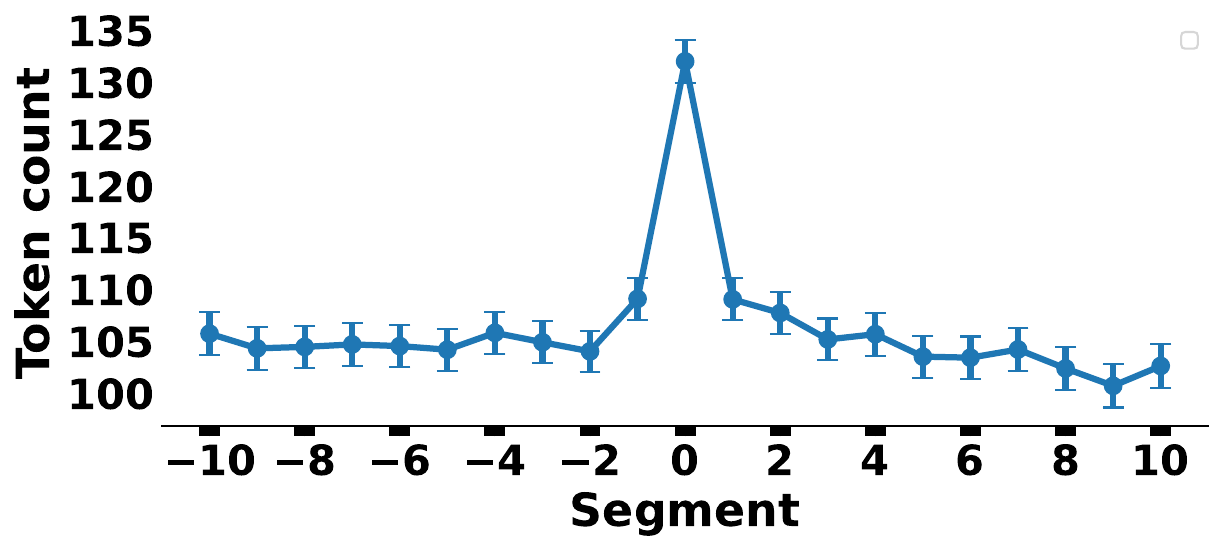}} & \textbf{\includegraphics[width=0.5\linewidth]{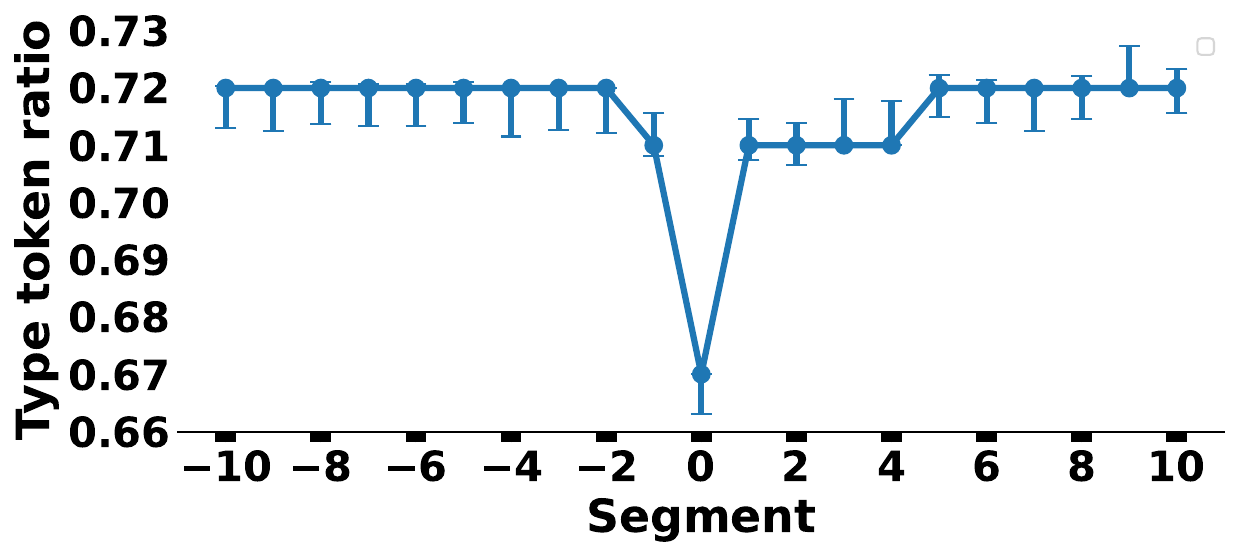}}
\end{tabular}
\captionof{figure}{\label{fig:more_textual_statistics}\footnotesize{Token count (left) and type token ratio (right) across segments at 95\% confidence intervals.}}
\end{table}

\subsection{Keywords}
\begin{figure}[!ht]
    \centering
    \includegraphics[width=1.0\linewidth]{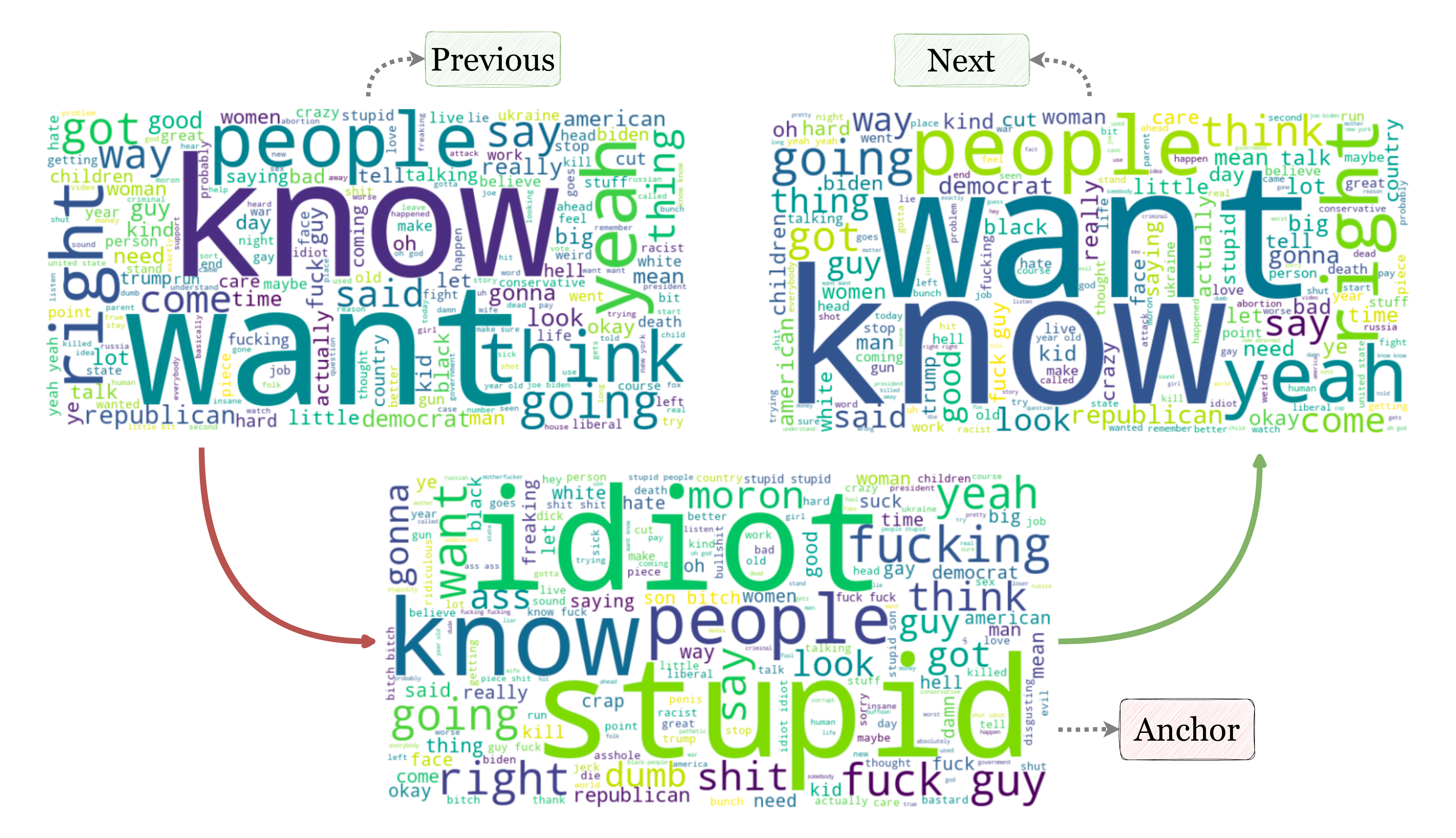}
    \caption{\footnotesize{\textsc{\textbf{Word Clouds}}: previous, anchor and next segments.}}
    \label{fig:word_cloud}
\end{figure}

To identify and compare the keywords present in the segments, we extract the top ten toxic key-phrases from each segment in every conversation chain using \textsc{KeyBERT}~\cite{grootendorst2020keybert}. Each phrase is limited to a maximum of five-grams. The word embeddings for \textsc{KeyBERT} algorithm are obtained from the \textsc{Detoxify}~\cite{Detoxify} library\footnote{\url{https://huggingface.co/unitary/toxic-bert}}. To ensure diversity in our extracted key-phrases, we use the maximal margin relevance utility of the \textsc{KeyBERT} algorithm, enforcing a diversity score of 0.75. In addition, we eliminate stop words \& punctuation and only retain strings containing characters from the English alphabet. We combine all the keywords obtained from the preceding ten segments and represent them as a word cloud; similarly, we combine all the keywords obtained from the following ten segments and represent them as another word cloud. We also obtain the word cloud for the anchor segment separately. These word clouds in series are illustrated in Figure~\ref{fig:word_cloud}. We observe that most of the words are linked to political and controversial themes like `right', `republican', `democrat', `women', `biden', `american' \& `liberal', which naturally follows from the choice of our dataset. The anchor segment has several toxic keywords including `idiot', `stupid', `f*cking/f*ck', `sh*t', `a*s' \& `moron'. The high similarity between the previous and next word clouds indicates that the anchor segment introduces a disruption in the flow of the main conversation, which rewinds back to normal only at the end of the anchor segment. Finally, in both the previous and the next word clouds, words like `want', `know', `people' \& `yeah' appear, which reflect an expression of demand. Thus, hostility in the speech of a particular speaker seems to be fueled by words of demand from either the anchor or other speakers/participants in podcast conversations.

\begin{table*}[!ht]
\centering
\setlength{\tabcolsep}{1.5mm}
\renewcommand{\arraystretch}{6}
\setlength{\tabcolsep}{0.001mm}
\begin{tabular}{lll}
\textbf{\includegraphics[width=0.33\linewidth]{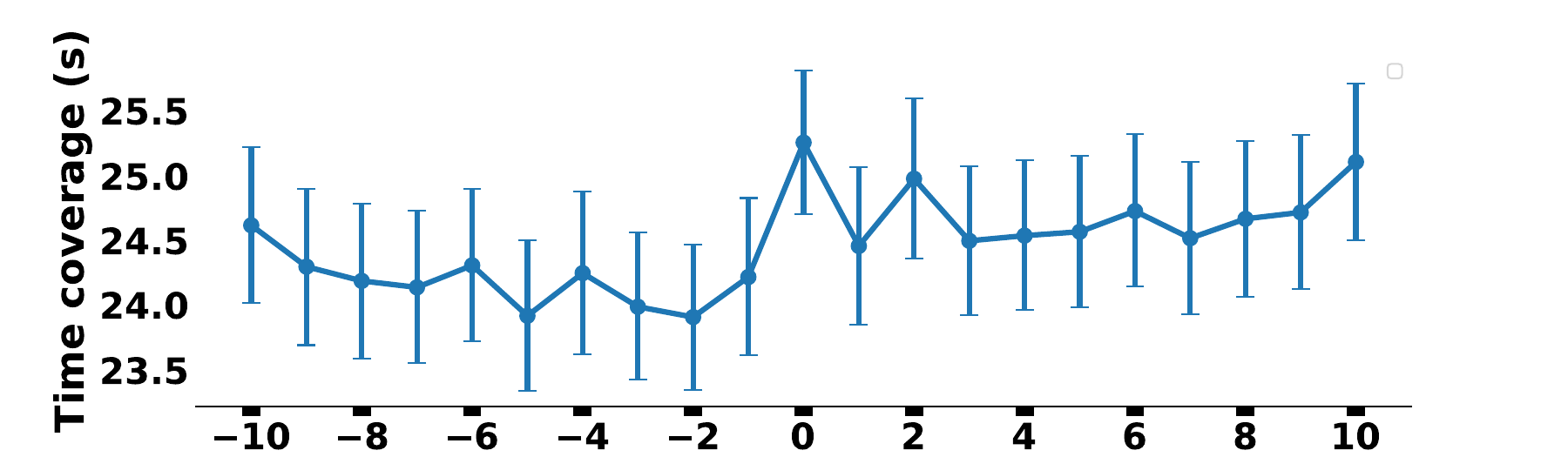}} & \textbf{\includegraphics[width=0.33\linewidth]{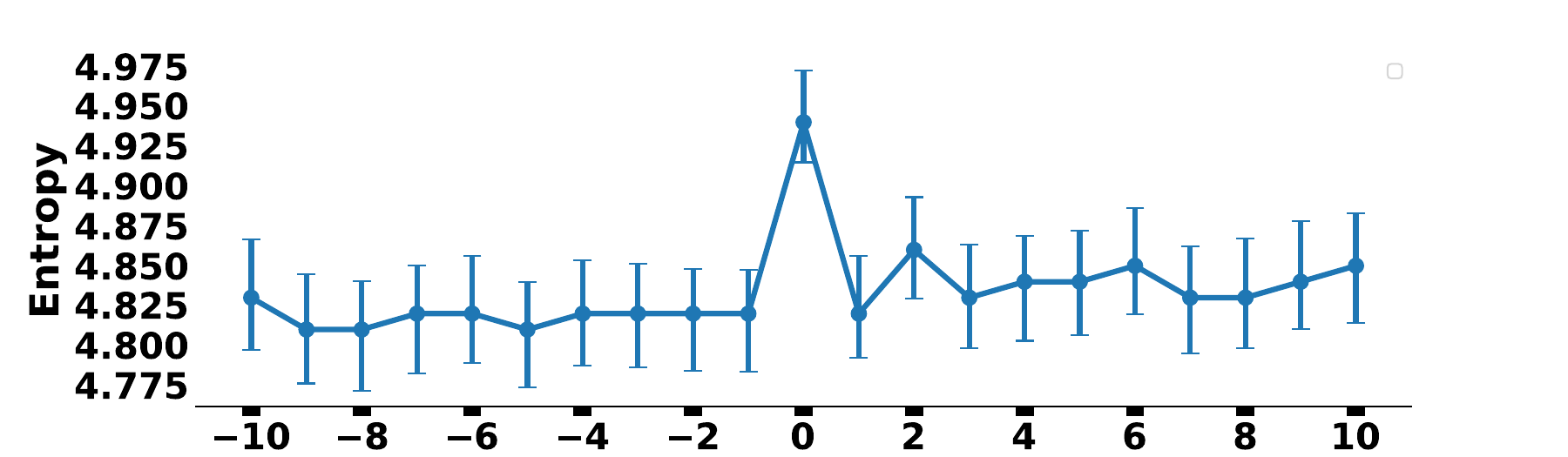}} &
\textbf{\includegraphics[width=0.34\linewidth]{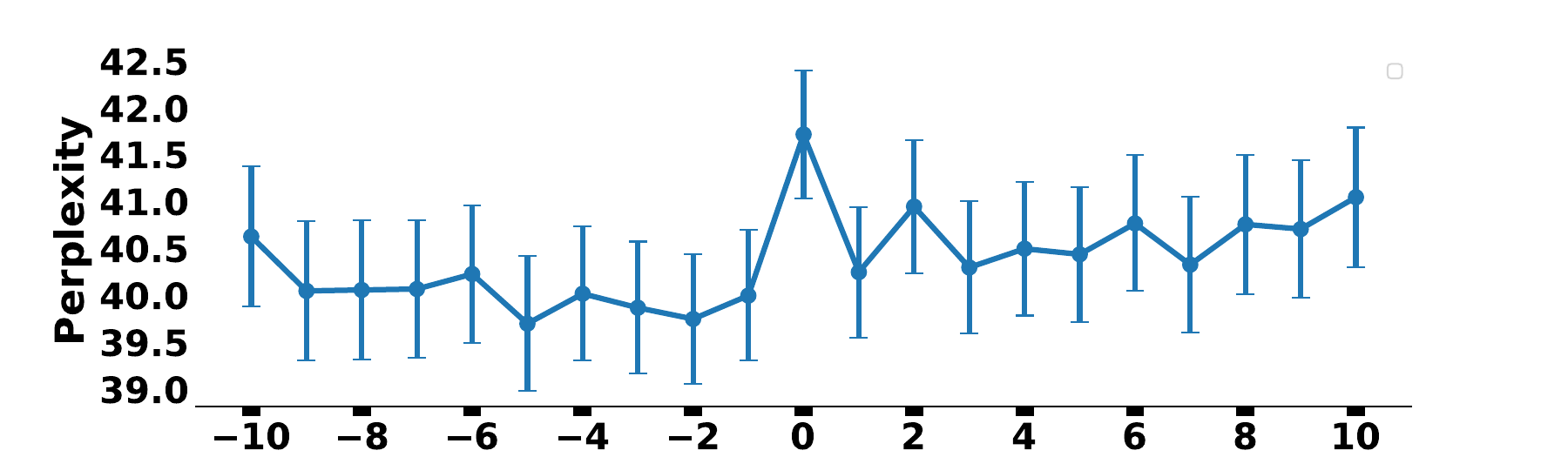}} \\ \textbf{\includegraphics[width=0.33\linewidth]{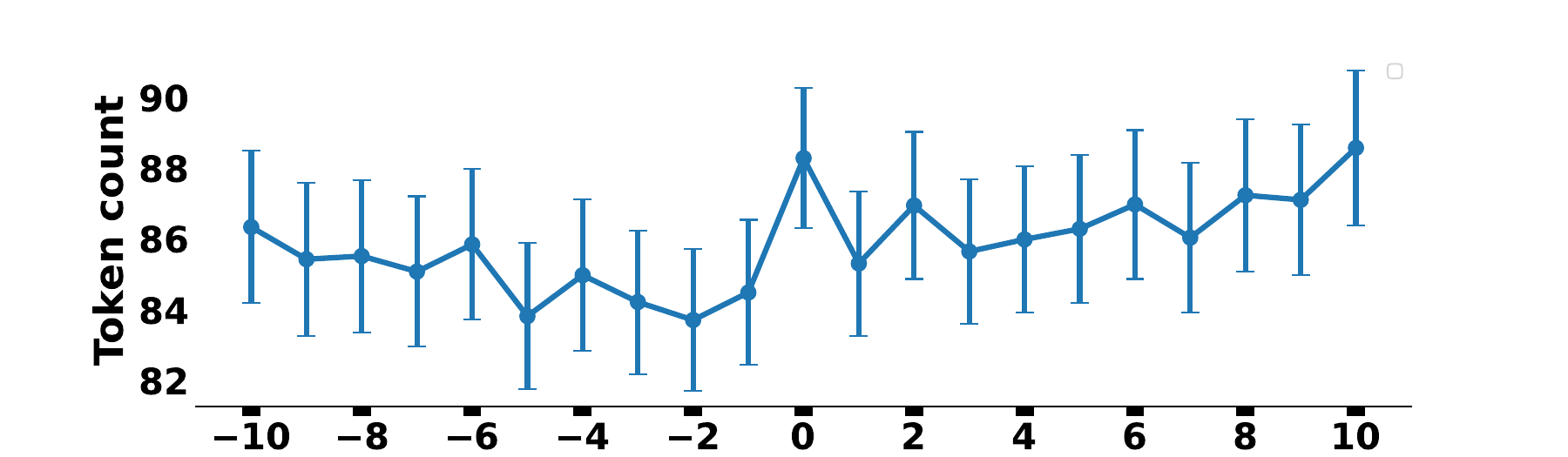}} &
\textbf{\includegraphics[width=0.33\linewidth]{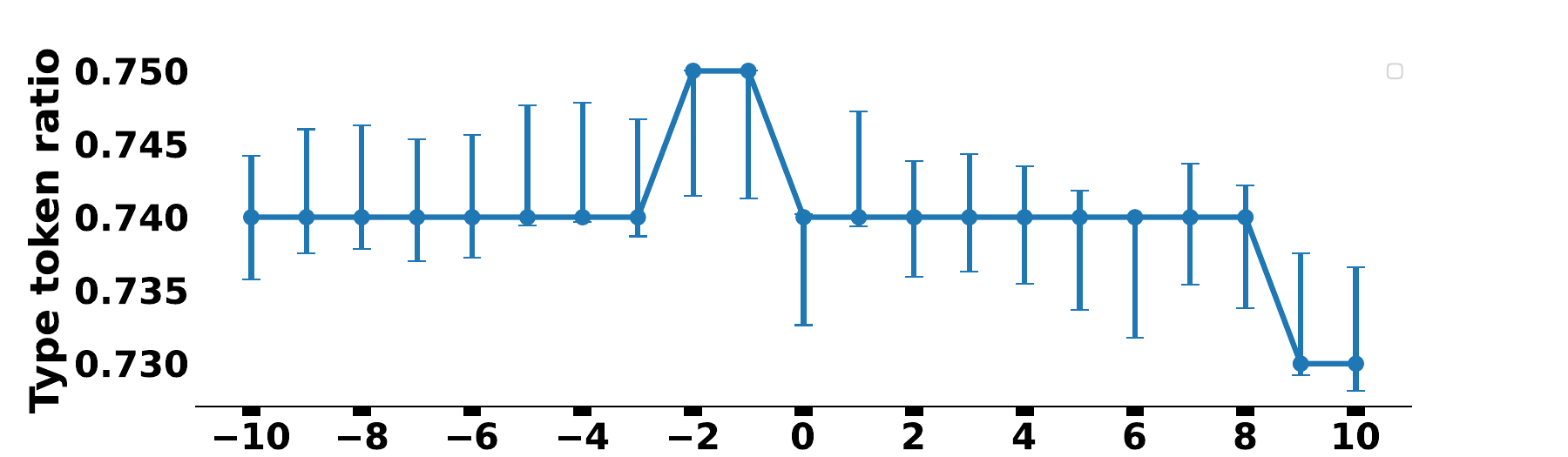}}
\end{tabular}
\captionof{figure}{\label{fig:control_statistics}\footnotesize{\textsc{\textbf{Control Group Statistics:}} Mean at 95\% confidence intervals for time coverage, entropy, perplexity, token count \& ttr for control group across segments.}}
\end{table*}

\begin{table*}[!ht]
\centering
\setlength{\tabcolsep}{1.5mm}
\renewcommand{\arraystretch}{6}
\setlength{\tabcolsep}{0.001mm}
\begin{tabular}{lll}
\textbf{\includegraphics[width=0.33\linewidth]{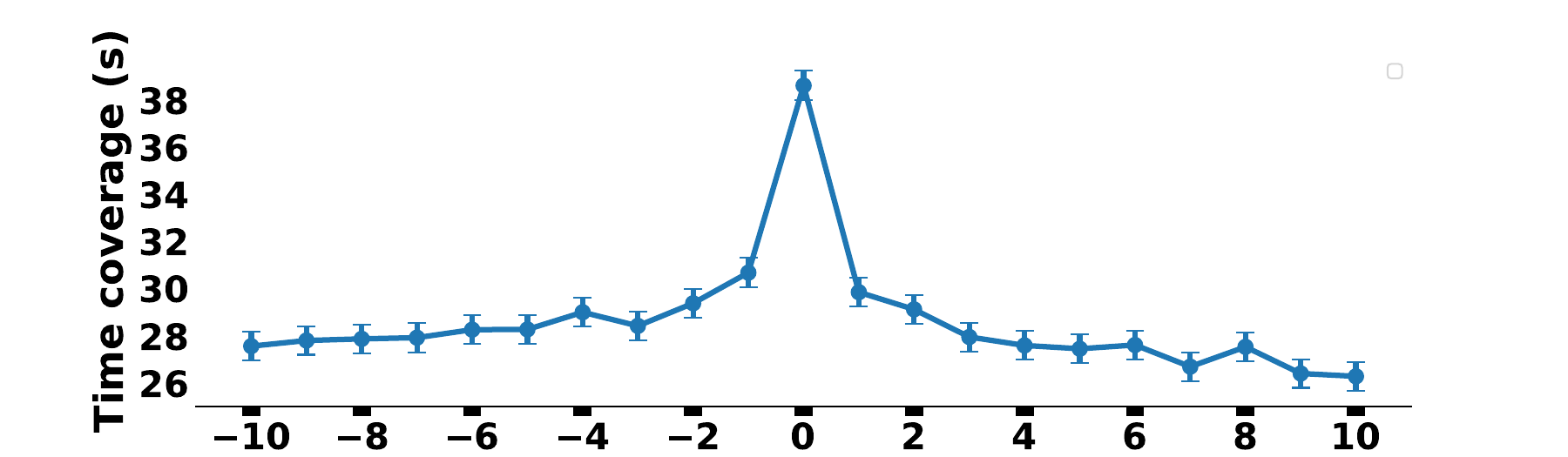}} & \textbf{\includegraphics[width=0.33\linewidth]{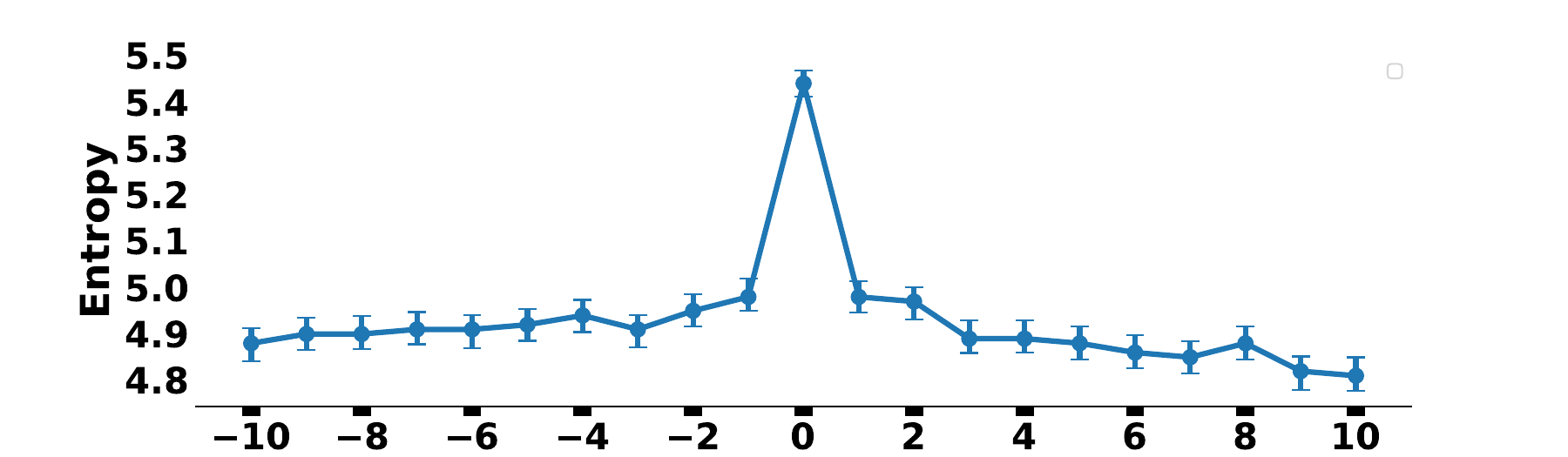}} &
\textbf{\includegraphics[width=0.34\linewidth]{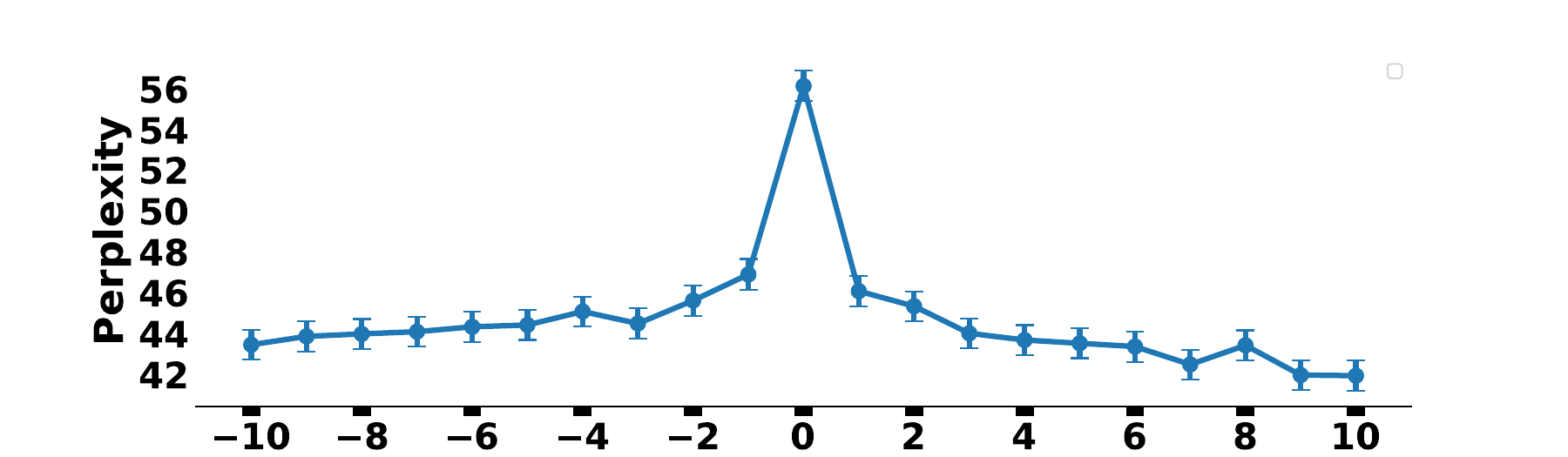}} \\ \textbf{\includegraphics[width=0.33\linewidth]{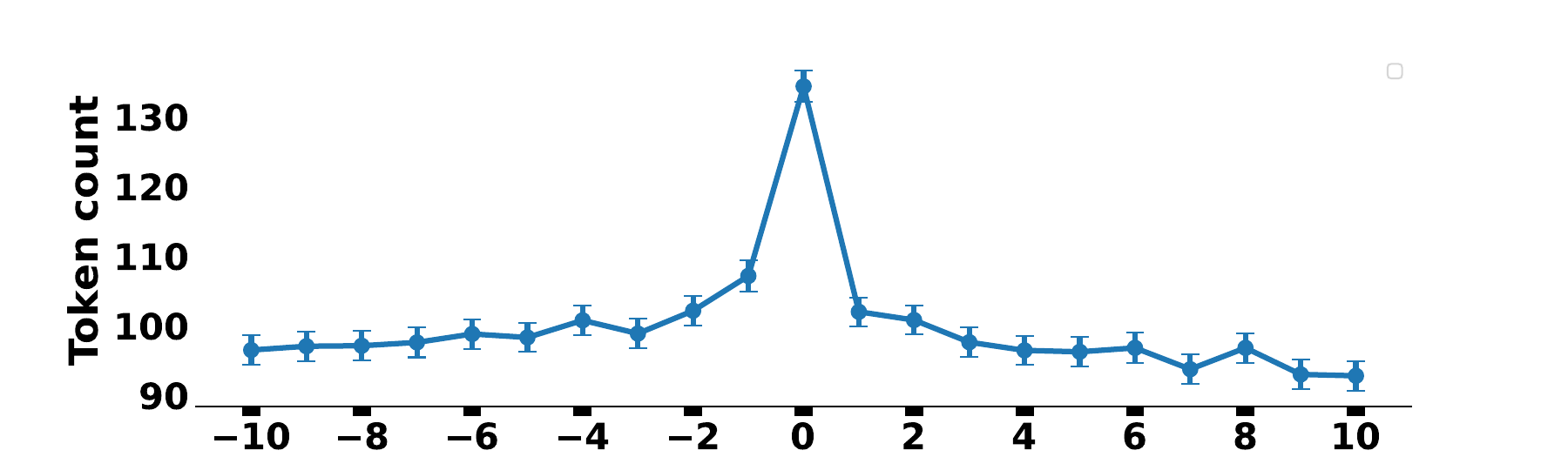}} &
\textbf{\includegraphics[width=0.33\linewidth]{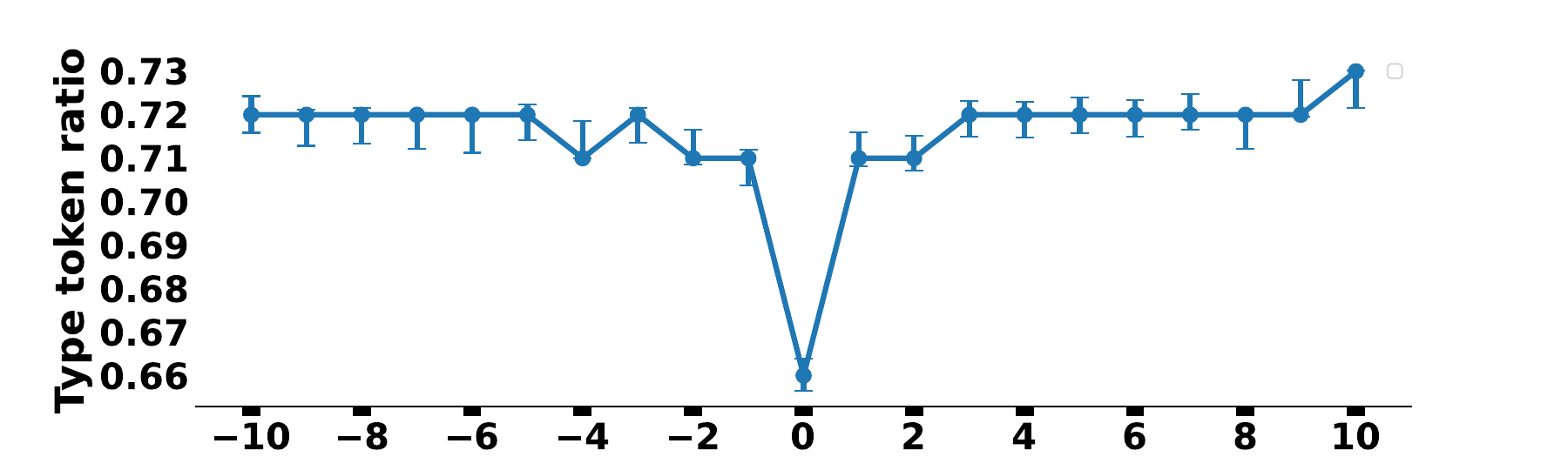}}
\end{tabular}
\captionof{figure}{\label{fig:left_statistics}\footnotesize{\textsc{\textbf{Left-leaning Statistics:}} Mean at 95\% confidence intervals for time coverage, entropy, perplexity, token count \& ttr for left-leaning podcast channels across segments.}}
\end{table*}

\begin{table}[!ht]
\footnotesize
\centering
\renewcommand{\arraystretch}{1.3}
\setlength{\tabcolsep}{1mm}
\caption{\footnotesize{Topic transitions in the conversation chains induced by \textbf{\textsc{BERTopic}}.}}
\begin{tabular}{c|c}
\textbf{Order in chain} & \textbf{Induced topics}                                         \\ \hline
\textbf{Preceding}               & {\textit{like, know, people, go, right, yeah, think, get, say, going}}     \\
\textbf{Anchor}                  & {\textit{b**ch, stupid, son, f**k, s*it, f**king, idiot, damn, shut, guy}} \\
\textbf{Following}               & {\textit{like, know, people, go, right, yeah, think, get, say, going}}    
\end{tabular}
\label{tab:topic-modeling}
\end{table}

\subsection{Topical shifts}
The keyword analysis in the previous section indicates that there is a significant change in the conversation content during the transition from the previous to the anchor segment. This hints at the fact that there is a possible topical/thematic shift during such a transition. In order to establish this, we perform topic modeling using the BERTopic~\cite{grootendorst2022bertopicneuraltopicmodeling} model. We extract the topics considering the previous ten aggregated segments as one document, the next ten aggregated segments as a second document and the anchor segment as the third document. We set the number of topics to three and report the top ten most representative words for that topic, which has the highest probability of association with a document. The results are noted in Table \ref{tab:topic-modeling}, which reveals significant shifts in thematic focus and toxicity level during the transition from the previous to the anchor segment and the anchor segment to the next segment. Precisely, the anchor topic is highly toxic while the preceding and the following topics are more related to demands, thus offering insights into the progression and contextual drivers of toxic conversations.

\subsection{Number of speakers in toxic chains}
We present the number of speaker turns and its distribution over 8,634 toxic chains from right-leaning channels presented as (\#speakers: \% toxic chain)-- {(1: 1.26), (2: 8.78), (3: 16.89), (4: 21.35), (5: 19.21), (6: 14.94), (7: 8.92), (8: 4.91), (9: 2.11), (10 and more: 1.63)}. Notably, only a small percentage of chains, i.e., 1.26\% have one speaker within them. Hence, we conclude that the toxic chains represent conversations rather than monologue behaviour.

\subsection{Right-leaning control group}

Here we cover the analysis of non-toxic chains where we choose such conversations that have the toxicity value of the anchor segment lower than 0.3 \textit{(as per the recommendation of Perspective API)} for right-leaning channels. We also take care that the randomly selected chains have a distribution in line with the distribution of podcast channels in toxic conversation chains. Further, we exclude such chains that have the previous and next ten segments with toxicity higher than 0.3 to ensure a consistent and fair comparison.\\
Results are presented in Figure~\ref{fig:control_statistics}. It is evident from the figure that, unlike toxic conversation, non-toxic conversation is more organized \& consistent with low randomness.

\end{document}